\vspace{-5pt}

\documentclass[10pt,twocolumn,letterpaper]{article}

\usepackage{cvpr}              

\usepackage[accsupp]{axessibility}  
\usepackage{CJKutf8}
\usepackage{float}
\usepackage{amsmath}
\usepackage{amssymb}
\usepackage{booktabs}
\usepackage{algorithm}
\usepackage{multirow}
\usepackage{algorithmic}
\usepackage{graphicx}
\usepackage{bm}
\usepackage{appendix}

\usepackage[pagebackref,breaklinks,colorlinks]{hyperref}

\usepackage[capitalize]{cleveref}
\crefname{section}{Sec.}{Secs.}
\Crefname{section}{Section}{Sections}
\Crefname{table}{Table}{Tables}
\crefname{table}{Tab.}{Tabs.}

\def\cvprPaperID{1514} 
\def\confName{CVPR}
\def\confYear{2022}

\begin{document}

\title{Object Localization under Single Coarse Point Supervision}


\author{Xuehui Yu$^{1}$\thanks{\ Equal contribution.}, Pengfei Chen$^{1*}$, Di Wu$^{1}$, Najmul Hassan$^{2}$, Guorong Li$^{1}$, \\ 
Junchi Yan$^{3}$, Humphrey Shi$^{2,4}$, Qixiang Ye$^{1}$, Zhenjun Han$^{1}$\thanks{\ Corresponding authors. (hanzhj@ucas.ac.cn)} \\
{\small \textsuperscript{1}University of Chinese Academy of Sciences, \textsuperscript{2}U of Oregon, \textsuperscript{3}Shanghai Jiao Tong University, \textsuperscript{4}Picsart AI Research (PAIR)}}
\maketitle

\begin{abstract}

Point-based object localization (POL), which pursues high-performance object sensing under low-cost data annotation, has attracted increased attention.
However, the point annotation mode inevitably introduces semantic variance for the inconsistency of annotated points. Existing POL methods heavily reply on accurate key-point annotations which are difficult to define.
In this study, we propose a POL method using coarse point annotations, relaxing the supervision signals from accurate key points to freely spotted points.
To this end, we propose a coarse point refinement (CPR) approach, 
which to our best knowledge is the first attempt to alleviate semantic variance from the perspective of algorithm.
CPR constructs point bags, selects semantic-correlated points, and produces semantic center points through multiple instance learning (MIL).
In this way, CPR defines a weakly supervised evolution procedure, which ensures training high-performance object localizer under coarse point supervision.
Experimental results on COCO, DOTA and our proposed SeaPerson dataset validate the effectiveness of the CPR approach. 
The dataset and code will be available at \color{magenta}\url{https://github.com/ucas-vg/PointTinyBenchmark/}.

\end{abstract}

\vspace{-2em}
\section{Introduction}
\vspace{-0.5em}
Humans can recognize and easily achieve a sense of the objects present in their eye-sight. In computer-vision, this is usually framed as drawing bounding boxes around objects ~\cite{DBLP:conf/iccv/LinGGHD17, DBLP:conf/nips/RenHGS15,yang2021cvpr,yang2020eccv} or dense annotations of the entire scene ~\cite{he2017mask, huang2019ccnet}. However,  one inevitable circumstance for training such models is that they require high-quality densely annotated data which is expensive and difficult to obtain. In some applications~\cite{runow2020deep}, just the object's location is necessary while costly annotation (e.g. bounding box) is redundant or even undesirable (e.g.\ a robotic arm aims at a single point to pick up an object~\cite{runow2020deep}). 

\begin{figure}[tb!]
\begin{center}
\setlength{\belowcaptionskip}{-1.0cm}
    \begin{tabular}{ccc}
    \includegraphics[height=0.44\linewidth]{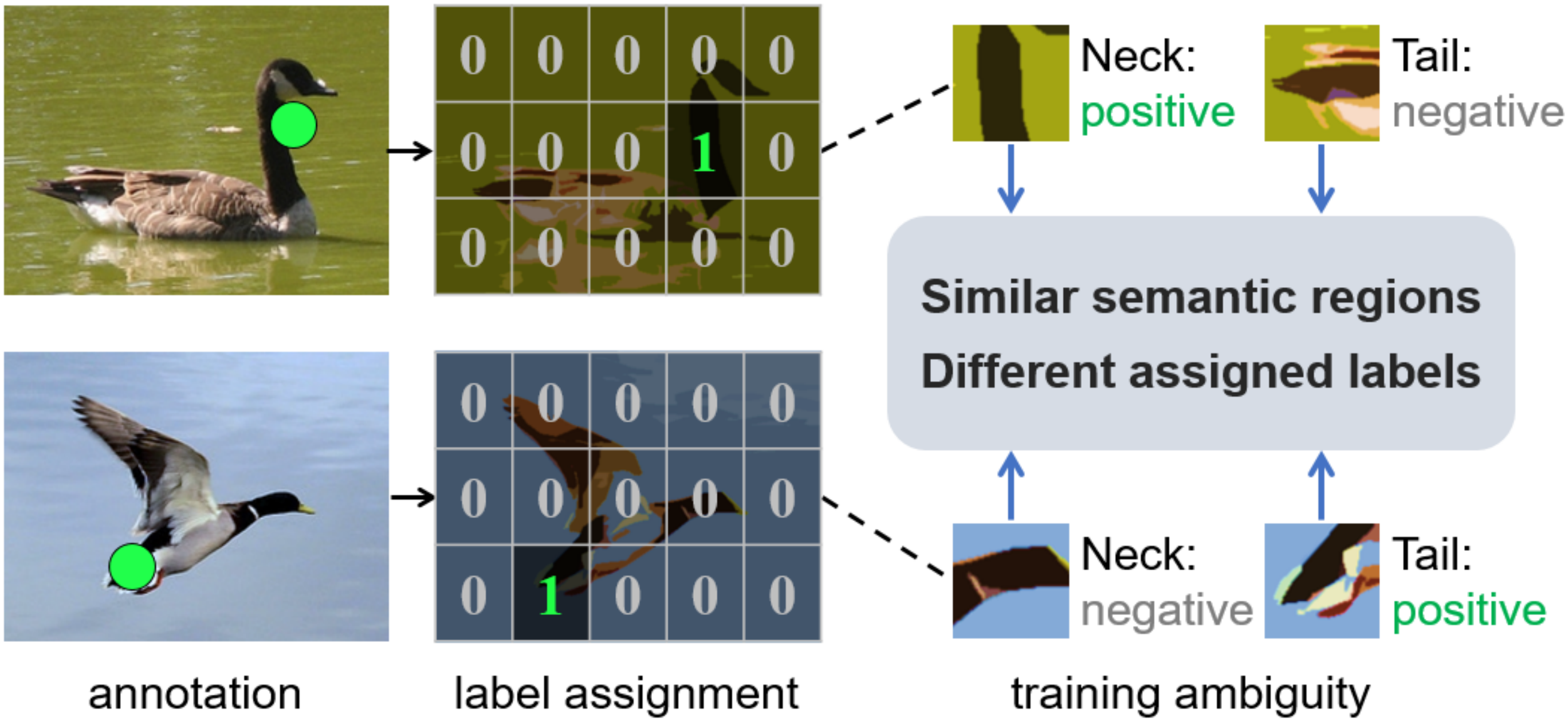}
    \end{tabular}
    \vspace{-15pt}
    \caption{Examples of coarse point annotation and the problem of semantic variance. (Best viewed in color.)
    }
\label{fig:sementic variance}
\end{center}
\end{figure}  

Hence point-based object localization (POL) is studied. Due to the simple and time-efficient annotation, point-based object localization has attracted increasing attention in recent years ~\cite{DBLP:conf/cvpr/RiberaGCD19, Song_2021_ICCV}. POL based methods require point-level object annotations and can predict the object's location as a 2D coordinate in the original image.

However, while annotating an object as a point, there can be multiple candidate points. One problem that arises with optional candidate points is that multiple regions of varying semantic information are labeled positive for the same class. Conversely, identical regions with similar semantic information are labeled differently. 
Take \textit{bird} category as an example, during annotation, we label the bird's different body parts (e.g.\ neck and tail \textit{etc.}) as positive based on the visible regions in the image. Based on the annotation, for different images in the dataset, we have labeled same body part (e.g.\ neck) of the bird as both positive and negative (see Fig.~\ref{fig:sementic variance}). Therefore, during training, the model has to consider the neck region as positive for one image and negative in another (the image where tail is annotated). This phenomenon introduces ambiguity and confuses the model which results in poor performance.

\begin{figure}[tb!]
  \centering
    \begin{subfigure}{0.495\linewidth}
    \centering
    \includegraphics[height=0.61\linewidth]{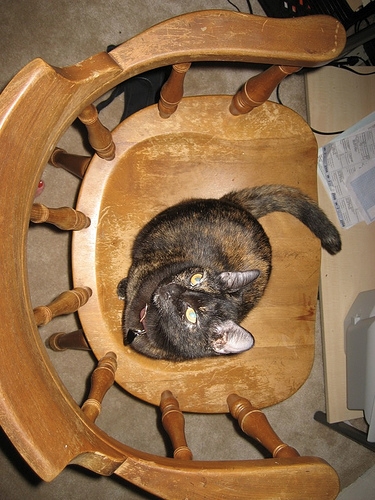}
    \includegraphics[height=0.61\linewidth]{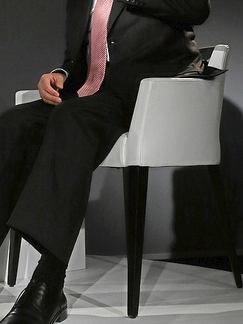}
    \caption{Chairs with different shapes.
    }
    \label{fig:short-b}
  \end{subfigure}
  \hfill
  \begin{subfigure}{0.495\linewidth}
    \centering
    \includegraphics[height=0.61\linewidth]{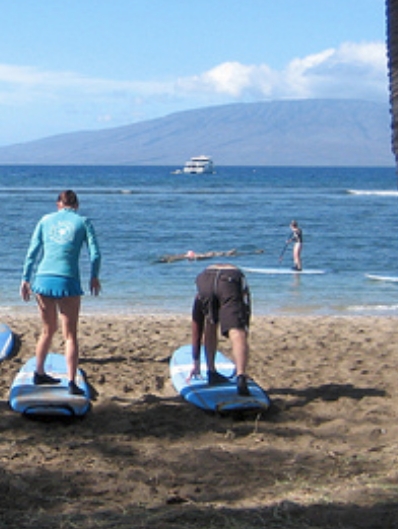}
    \includegraphics[height=0.61\linewidth]{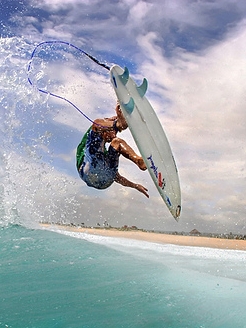}
    \caption{Persons with different poses.}
    \label{fig:short-b}
  \end{subfigure}
  \hfill
  \begin{subfigure}{1.0\linewidth}
    \centering
    \includegraphics[height=0.31\linewidth]{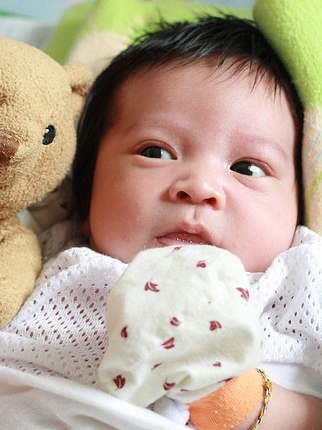}
    \includegraphics[height=0.31\linewidth]{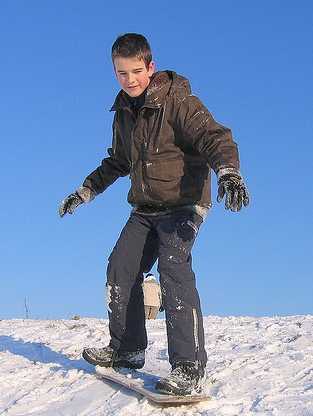}
    \includegraphics[height=0.31\linewidth]{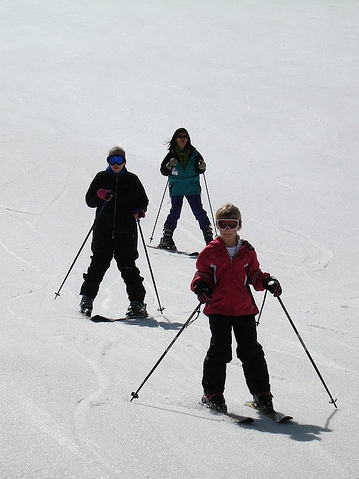}
    \includegraphics[height=0.31\linewidth]{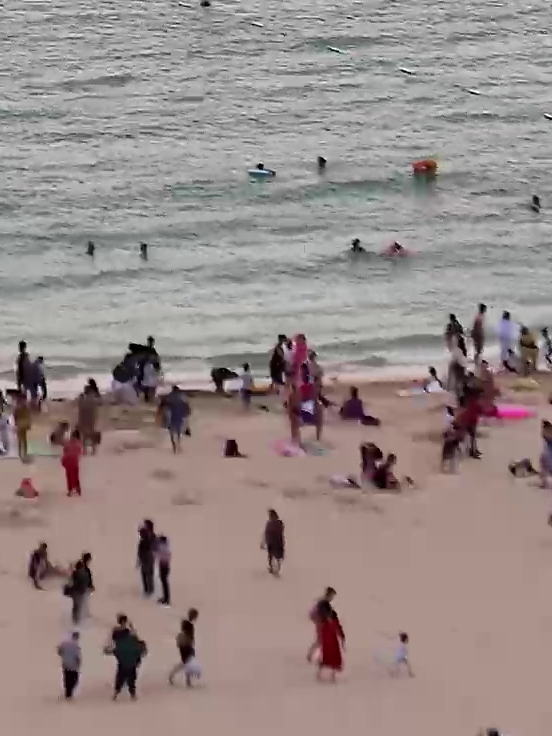}
    \caption{Persons of different sizes.}
    \label{fig:short-b}
  \end{subfigure}
\vspace{-10pt}
\setlength{\belowcaptionskip}{-0.5cm}
  \caption{The difficulties of key-point based annotation. (a) Key points are hard to define due to the large in-class variance of shape. (b) Key point (e.g.\ head) does not exist due to multiple poses and views. (c) Key point's granularity (eye, forehead, head or body) is hard to determine due to multiple scales.}
\label{fig:key point problem}
\end{figure}

Previous works~\cite{Song_2021_ICCV,DBLP:journals/pami/WangGLL21} addressed this issue by setting strict annotation rules by annotating only the pre-defined key-point areas of the object. As a result, they suffer from the following challenges:
\textbf{i)} the key points are not easy to define, especially for some broadly defined categories where they do not have a specific shape (Fig.~\ref{fig:key point problem} (a));
\textbf{ii)} the key point may not exist in the image due to the different poses of objects and different camera views (Fig.~\ref{fig:key point problem} (b));
\textbf{iii)} when objects have large scale variance, it is difficult to decide the appropriate granularity of the key points ( Fig.~\ref{fig:key point problem} (c)). For a person, if the head is a key point~\cite{Song_2021_ICCV} (coarse-grained) then there remains a large semantic variance for the large-scale instance (whether to annotate the eye or nose). If the eye is labeled as a key point~\cite{DBLP:journals/pami/WangGLL21} (fine-grained) then the position of eyes for a small-scale instance cannot be identified. Thus, the complicated annotation rules are required to solve the semantic variance problem from annotation perspective, which considerably increases the annotation difficulty and human burden.
Therefore, the challenges mentioned above restrict previous POL methods from exploring multi-class and multi-scale datasets (e.g.\ COCO or DOTA).

In this paper, we formulate the coarse point-based localization (CPL) paradigm for training a localizer of a general POL, as shown in Fig.~\ref{fig:pipeline}. We firstly adopt a coarse point annotation stragegy, which allows to annotate any point on an object. Then the coarse point refinement (\textbf{CPR}) algorithm is proposed to refine the initialized annotated coarse point to the semantic center in the training set. Finally, the refined points instead of the annotated points are used as supervision to train a localizer. The proposed CPR is the first attempt to alleviate semantic variance from the perspective of algorithm rather than annotation. 
Specifically, CPR finds the semantic points around the annotated point through multiple instance learning (MIL) ~\cite{DBLP:journals/ai/DietterichLL97}, then weighted averages the semantic points to obtain the semantic center, which has a smaller semantic variance and a higher tolerance for prediction errors. The contributions are:

1) We dive into point-based object localization (POL) task, and formulate the coarse point based localization (CPL) paradigm for general object localization, extending the previous works to a multi-class/multi-scale POL task;

2) The coarse point refinement (CPR) algorithm is proposed to alleviate the semantic variance from the perspective of algorithm rather than rigid annotation rules;

3) The experimental results show the CPR is effective for CPL, which obtains a comparable performance with the center point (approximate key point) based object localization, and improves the performance over 10 points compared with the baseline;

4) A new dataset with more than 600,000 annotations, named SeaPerson, is introduced in this paper. This dataset can be used for tiny person detection and localization. 

\section{Related Work}

In this section, we review the relevant point-based vision tasks and the vision tasks with multiple instance learning. 

\subsection{Vision Tasks under Point Supervision} 
\textbf{Pose Estimation.} 
Human or animal pose estimation aims to locate the position of joint points of persons or animals accurately~\cite{sun2019deep, zhang2021towards, zhang2022vitaev2}. 
There are several benchmarks built for the task, e.g.\ COCO ~\cite{lin2014microsoft} and the Human3.6M ~\cite{DBLP:journals/pami/IonescuPOS14} datasets are the most well-known ones for 2D and 3D pose estimation and AP-10k~\cite{yu2021ap} for animal pose estimation.
In these datasets, annotations are a set of accurate key points, and the predicted results are human or aninimal poses rather than the location of person or animal instances.

\textbf{Crowd Counting.} In this task, accurate head annotation is utilized as point supervision~\cite{DBLP:conf/cvpr/ZhangZCGM16, DBLP:journals/pami/WangGLL21, Song_2021_ICCV}. The crowd density map~\cite{DBLP:conf/nips/LempitskyZ10, DBLP:conf/cvpr/JiangZXZLZYP20,DBLP:conf/eccv/HuJLZHCD20}, generated by head annotation, is chosen as the optimization objective of the network. Furthermore, crowd counting focuses on the number of people rather than each person's position. It depends on precise key points such as the human head, while the coarse point object localization task only requires the coarse position annotation on the human body.
 
\textbf{Object Localization.} 
Unlike object detection~\cite{DBLP:conf/iccv/LinGGHD17, DBLP:conf/nips/RenHGS15, DBLP:fusionfactor}, especially for rotation detection~\cite{yang2021icml,yang2021nips}, requiring the exact bounding box information, object localization applications ~\cite{DBLP:conf/cvpr/RiberaGCD19} are often agnostic to object's scale. The works~\cite{DBLP:conf/cvpr/RiberaGCD19, Song_2021_ICCV} train a localizer 
with points instead of bounding boxes. These tasks are summarized as POL in our paper. However, they heavily rely on key-point annotations to reduce the semantic variance.

Different from the above mentioned tasks, our CPL relies upon a coarse point instead of keypoints and deals the semantic variance problem with a novel approach.
\subsection{Vision Tasks with Multiple Instance Learning}
The paradigm of MIL~\cite{DBLP:journals/ai/DietterichLL97} is that a bag is positively labeled if it contains at least one positive instance; otherwise, it is labeled as a negative bag. 
Inspired by weakly supervised object detection task, the proposed CPR method follows the MIL paradigm. With the object category and the coarse point annotation, we consider sampled points around each annotated point as a bag and utilize MIL for training. 

\textbf{Image-level Tasks.} An image is divided into patches, where patches are seemed as instances and the entire image as a bag. Content-based image retrieval ~\cite{DBLP:journals/pr/ZhangWSZ10,DBLP:conf/icml/ZhangGYF02} is a conventional MIL task, which just classifys images by their content. 
If the image contains at least one object of a class, the whole bag can be seen as a positive sample for that class.
Otherwise, the bag will be regarded as a negative sample.

\textbf{Video-level Tasks.} Firstly, the video is divided into segments, which will be classified separately and then the whole video is seemed as a bag. Following the above pre-processing, MIL is used to identify specific events in videos~\cite{DBLP:conf/cvpr/FengHZ21,DBLP:conf/cvpr/NguyenLPH18,DBLP:conf/cvpr/SultaniCS18}. Additionally, some researchers have applied MIL to video object tracking~\cite{DBLP:conf/cvpr/BabenkoYB09,DBLP:journals/pami/BabenkoYB11}. 
\cite{DBLP:journals/pami/BabenkoYB11} also achieves a robust tracker by constructing instance-level bag from box, but different from our work, it does not handle the negative samples to suppress background for MIL.

\textbf{Object-level Tasks.}
MIL is widely used in weakly supervised object localization and detection (WSOL~\cite{DBLP:journals/pami/CinbisVS17, wu2021background} and WSOD~\cite{DBLP:conf/cvpr/BilenV16, DBLP:conf/ijcai/WangYZZ18,DBLP:conf/cvpr/WanLKJJY19, DBLP:journals/pami/TangWBSBLY20, DBLP:conf/cvpr/ChenFJC020, DBLP:journals/pami/WanWHJY19}), where only the image-level annotation is utilized. Firstly, Select Search~\cite{DBLP:conf/iccv/SandeUGS11} or Edge Box~\cite{DBLP:conf/eccv/ZitnickD14} methods are used to produce proposal boxes, which are then used as a bag and each of them as an instance. 
Finally, they classified positive and negative samples by judging whether the image contains at least one object of a specific class. WSOL/WSOD, only with image-level annotation, focus on local regions and can not distinguish instances due to the lack of object-level annotation. 

Annotation of CPL is a coarse point position and the category of each object. CPR views sampled points around the annotated point as a bag and trains object-level MIL to find a better and stable semantic center.

\begin{figure}[tb!]
\begin{center}
    \begin{tabular}{ccc}
    \includegraphics[width=0.90\linewidth]{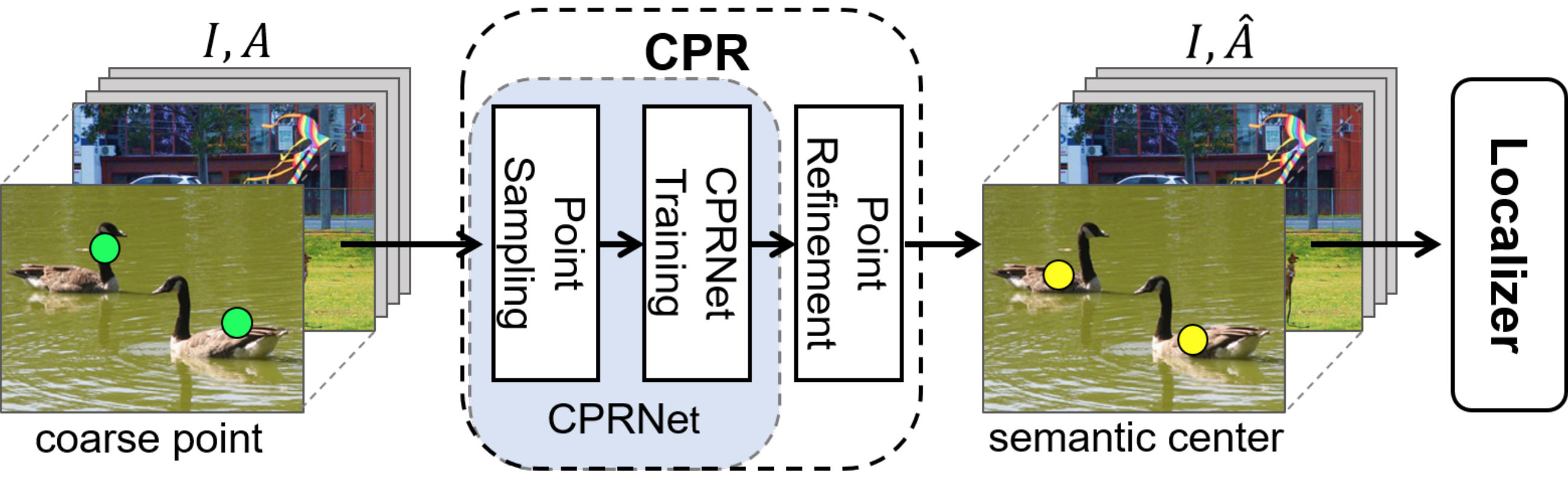}
    \end{tabular}
\vspace{-15pt}
\setlength{\belowcaptionskip}{-0.8cm}
   \caption{Pipeline of CPL in three steps: 1) Annotating objects as coarse points $A$. 2) Refining annotated points to semantic centers $\hat{A}$. 3) Training a localizer (e.g. P2PNet) with $\hat{A}$ as supervision.}
\label{fig:pipeline}
\end{center}
\end{figure}

\begin{figure*}[tb!]
\begin{center}
    \begin{tabular}{ccc}
    \includegraphics[width=0.98\linewidth]{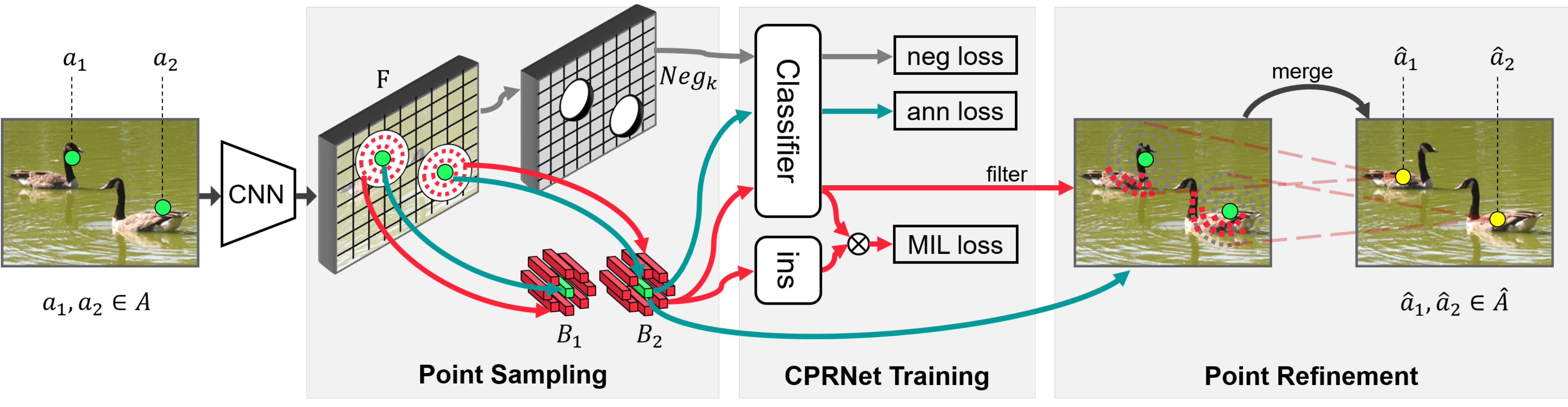}
    \end{tabular}
\vspace{-15pt}
\setlength{\belowcaptionskip}{-0.8cm}
   \caption{The framework of CPR. With $F$, there are three steps in CPR:
   1) Points bag (e.g. $B_1, B_2$) and negative samples (e.g. $Neg_k$) are obtained by point sampling according to annotated points $A$ (green), and then feature vectors of these points are extracted on $F$. 2) CPRNet are trained with the feature vectors based on MIL loss, annotation loss and negative loss. 3)
   Semantic points (red points on the birds) are selected by classification scores of points in bag (e.g. $B_1, B_2$) predicted by the trained CPRNet. Finally, the refined points (yellow) $\hat{A}$ are obtained by weighted averaging the semantic points. (Best viewed in color.)
   }
\label{fig:framework}
\end{center}
\end{figure*}

\section{Coarse Point Refinement}
As shown in Fig.~\ref{fig:pipeline}, CPR can be regarded as a pre-processing that transforms the annotations on the training set to a more conductive form for the subsequent tasks. The main purpose of CPR is to find a semantic point, which has a smaller semantic variance and a higher tolerance for prediction errors, to replace the initial annotated point. 

In Fig~\ref{fig:framework} and Algorithm~\ref{Alg: coarse point refinement (CPR)}, there are three key steps in CPR: 
i) Point Sampling: points in the neighborhood of the annotated point are sampled; ii) CPRNet Training: based on the sampled points, a network is trained to classify whether the points are in the same category with the annotated point or not; iii) Point Refinement: 
based on the scores obtained by CPRNet and the constraints (details in Sec.~\ref{sec: Refinement}), the points, having similar semantic information with the annotated point, are chosen as the semantic points and then weighted with their scores to obtain the semantic center as the final refined point. 

\begin{algorithm}[tb!]
{
\small
\caption{Coarse Point Refinement}
\label{Alg: coarse point refinement (CPR)}
\textbf{Input:} Training set $D_{train}$, CPRNet $E$. \\
\textbf{Output:} Refined points $\hat{A}^{D_{train}}$.\\
\textbf{Note:} $A$ and $C$ are 2D coordinates and category label of annotated points in image $I$ respectively. \\
\vspace{-10pt}
\begin{algorithmic}[1]
\STATE $L_{CPR}^{D_{train}} \leftarrow 0$;
\FOR{$(I, A, C) \in D_{train}$}
\STATE Extract feature map $\mathbf{F}$ of $I$ with $E$;
\STATE \textit{// Step1: point sampling}
\STATE $B_j \leftarrow bag\_sampling(a_j)$ for each $a_j \in A$, Eq.~\ref{Eq:Bag_j};
\STATE $Neg_k \leftarrow neg\_sampling(k)$ for each category $k \in \{1,$ $2, ...K\}$, Eq.~\ref{Eq:Neg_k};
\STATE \textit{// Step2: CPRNet training}
\STATE Calculate $L_{MIL}$ with $B_j$ and $\mathbf{F}$, Eq.~\ref{Eq:L_{bag}};
\STATE Calculate $L_{ann}$ with $A$ and $\mathbf{F}$, Eq.~\ref{Eq:S_{a_j}};
\STATE Calculate $L_{neg}$ with $Neg_k$ and $\mathbf{F}$, Eq.~\ref{Eq:S_{Neg}};
\STATE Sum $L_{MIL}$, $L_{ann}$ and $L_{neg}$ to obtain the CPR loss $L_{CPR}$, Eq.~\ref{Eq:CPR basic loss};
\STATE $L_{CPR}^{D_{train}} \leftarrow L_{CPR}^{D_{train}} + L_{CPR}$;
\ENDFOR
\STATE Train $E$ by minimizing $L_{CPR}^{D_{train}}$ to obtain $\hat{E}$.
\STATE \textit{// Step3: point refinement}
\STATE $\hat{A}^{D_{train}} \leftarrow \{\}$;
\FOR{$(I, A, C) \in D_{train}$}
\STATE $\hat{A} \leftarrow Point\_Refinement(\hat{E}, I, A, C)$, Algorithm~\ref{Alg: refinement};
\STATE $\hat{A}^{D_{train}} \leftarrow \hat{A}^{D_{train}} \cup \{\hat{A}\}$;
\ENDFOR
\end{algorithmic}
}
\end{algorithm}


\subsection{Point Sampling}\label{sec:Point sampler}
In this paper, $K$ denotes the number of categories, $a_j\in R^2$ and $c_j \in\{0, 1\}^K$ denote the annotated point's 2D coordinate and the category label of $j$-th instance. $p=(p_x, p_y)$ denotes a point on a feature map.

\textbf{Point Bag Construction.} In Fig.~\ref{fig:framework}, to sample points uniformly in the neighborhood of $a_j$, we define $R$ circles with $a_j$ as the center, where the radius of the $r$-th ($1 \leq r \leq R$, $r\in N^+$) circle is set as $r$. Then we sample $r*u_0$ ($u_o$=8 by default) points with equal intervals around the circumference of the $r$-th circle, and obtain $Circle(a_j, r)$. All sampled points of the $R$ circles are defined as points' bag of $a_j$, denoted as $B_j$ in Eq.~\ref{Eq:Bag_j}. The points outside the feature map are excluded.
\begin{equation}\small
\begin{aligned}
Circle&(p, r)  = \bigg\{\bigg(p_x+r\cdot cos\left(2\pi \cdot \frac{i}{u_0 \cdot r}\right), p_y + \\
& r \cdot sin\left(2\pi\cdot\frac{i}{u_0\cdot r}\right)\bigg)\ |\  0 \leq i< r\cdot u_0, i\in N^+\bigg\}; \\
& B_j = \mathop{\cup}\limits_{1\leq r\leq R}\ Circle(a_j, r).
\label{Eq:Bag_j}
\end{aligned}
\end{equation}

$B_j$ is used for calculating the MIL loss for CPRNet training and obtaining the semantic points for point refinement.

\textbf{Negative Point Sampling.} 
All integer points on the feature map, outside the circles with radius $R$ of all annotated points of a given category, will be selected as negative samples. The negative samples of category $k$ can be defined as:
\vspace{-5pt}
\begin{equation}\small
\begin{aligned}
Neg_k & = \{(p_x, p_y)| p_x\leq w, p_y \leq h, p_x \in N^+, p_y \in N^+\\
& \forall (a_j, c_j)\ s.t. \ c_{jk}=1, ||p-a_j|| > R\},
\label{Eq:Neg_k}
\end{aligned}
\end{equation}
where $||p-a_j||$ is the Euclidean distance between $p$ and $a_j$. $w$ and $h$ are the width and height of a given feature map.

\subsection{CPRNet Training}
This section gives the details of the objective function of training CPRNet based on the sampled points bag $B_j$ ($j \in \{1, 2, ..M\}$) and the negative points $Neg_k$ ($k \in \{1, 2, ..K\}$), where $M$ and $K$ are the amount of instances and categories. $U$ is defined as the amount of points in $B_j$.

\textbf{CPRNet.}\label{sec:CPRNet architecture} CPRNet adopts FPN~\cite{DBLP:conf/cvpr/LinDGHHB17} with ResNet~\cite{DBLP:conf/cvpr/HeZRS16} as the backbone. Only P2 or P3 is used due to the lack of scale information in point annotation. After four 3$\times$3 conv layers followed by the ReLU ~\cite{DBLP:journals/jmlr/GlorotBB11} activation, the final feature map $\mathbf{F}\in \mathbb{R}^{h\times w \times d}$ is obtained, where $h\times w$ is 
the corresponding spatial size and $d$ is the dimension of channel. For a given point $p=(p_x, p_y)$, $\mathbf{F}_{p}\in \mathbb{R}^{d}$ denotes the feature vector of $p$ on $\mathbf{F}$. If $p$ is not an integer point, the bilinear interpolation is used to obtain $\mathbf{F}_p$. 

\textbf{CPR Loss.} Object-level MIL loss is introduced to endrow CPRNet the ability of finding semantic points around each annotated point. Then to overcome the over-fitting problem of MIL when the data is insufficient, we further introduce the instance-level prior as supervision by designing annotation and negative loss. The objective function of CPRNet is a weighted summation of the three losses:
\vspace{-2pt}
\begin{equation}\small
\begin{aligned}
L_{CPR} = & L_{MIL} + \alpha_{ann} L_{ann} +\alpha_{neg} L_{neg},
\label{Eq:CPR basic loss}
\end{aligned}
\end{equation}
\vspace{-2pt}
where $\alpha_{ann}=0.5$ and $\alpha_{neg}=3$ (by default in this paper). And $L_{MIL}$, $L_{ann}$ and $L_{neg}$ are based on the focal loss~\cite{DBLP:conf/iccv/LinGGHD17}:
\vspace{-6pt}
\begin{equation}\small
\begin{aligned}
FL(S_{p}, c_{j}) = \sum\limits_{k=1}^{K} c_{j,k}& (1 - S_{p, k})^{\gamma} \log(S_{p, k}) + \\
&(1-c_{j,k}) S_{p, k}^{\gamma} \log(1-S_{p, k}),
\label{Eq:focal loss}
\end{aligned}
\end{equation}
\vspace{-2pt}

\noindent where $\gamma$ is set as 2 in Eq.~\ref{Eq:focal loss}
following the standard focal loss, and $S_p \in \mathbb{R}^{K}$ and $c_{j} \in \{0, 1\}^{K}$ are the predicted scores on all categories and the category label, respectively.

\textbf{Object-level MIL Loss.} To find the semantic points during refinement, we refer to WSOD ~\cite{DBLP:conf/cvpr/BilenV16} and design a MIL loss to enable the CPRNet justify whether the points in $B_j$ are in the same category with $a_j$. 
Based on $B_j$, the feature vectors $\{\mathbf{F}_p|p\in B_j\}$ are extracted. As Eq.~\ref{Eq:S_{bag}} shows, for each $p \in B_j$, a classification branch $fc^{cls}$ is applied to obtain the logits $[\mathbf{O}^{cls}_{B_j}]_p$, which is then utilized as an input of an activation function $\sigma_1$ to obtain $[\mathbf{S}^{cls}_{B_j}]_p$. Besides, an instance selection branch $fc^{ins}$ is applied to $\mathbf{F}_p$ to obtain $[O^{ins}_{B_j}]_p$, which is then utilized as an input of an activation function $\sigma_2$ to obtain the selection score $[S^{ins}_{B_j}]_p$. The score $[S^{over}_{B_j}]_p$ is obtained by taking the element-wise product of $[S^{ins}_{B_j}]_p$ and $[S^{cls}_{B_j}]_p$.
\vspace{-2pt}
\begin{equation}\small
\begin{aligned}
&{[O^{cls}_{B_j}]}_p = fc^{cls}(F_p) \in \mathbb{R}^{K}, \quad
[O^{ins}_{B_j}]_p = fc^{ins}(F_p) \in \mathbb{R}^{K}; \\
&[S^{cls}_{B_j}]_p = [\sigma_1(O^{cls}_{B_j})]_p = 1/(1+e^{-[O^{cls}_{B_j}]_p})\in \mathbb{R}^{K}; \\
&[S^{ins}_{B_j}]_p = [\sigma_2(O^{ins}_{B_j})]_p = e^{O^{ins}_p}/\sum_{p' \in B_j} [e^{O^{ins}_{B_j}}]_{p'} \in \mathbb{R}^{K}; \\
&[S^{over}_{B_j}]_p = [S^{ins}_{B_j}]_p \cdot [S^{cls}_{B_j}]_p \in \mathbb{R}^{K},
\label{Eq:S_{bag}}
\end{aligned}
\end{equation}
\vspace{-8pt}

\noindent where $\sigma_2$ is a $softmax$ function. Different from MIL in WSOD, the $sigmoid$ activation function is applied for $\sigma_1$, due to its suitability for binary task compared with the $softmax$ function. Furthermore, the $sigmoid$ activation function allows to perform multi-label classification (for the overlapping area of multiply objects' neighborhood) for points and is more compatible with focal loss.

The bag-level score $S_{B_j}$ is obtained by the summation of all points' scores in $B_j$ by Eq.~\ref{Eq:S_{bag}2}. $S_{B_j}$ can be seen as the weighted summation of the classification score $[S^{cls}_{B_j}]_p$ of $p$ in $B_j$ by the corresponding selection score $[S^{ins}_{B_j}]_p$. 
\begin{equation}\small
\begin{aligned}
S_{B_j} & = \sum_{p \in B_j} [S^{over}_{B_j}]_p \in \mathbb{R}^{K}.
\label{Eq:S_{bag}2}
\end{aligned}
\end{equation}
The MIL loss is finally given by the focal loss on the predicted bag-level scores $S_{B_j}$ and the category label $c_j$ of $a_j$:
\vspace{-7pt}
\begin{equation}\small
\begin{aligned}
L_{MIL} & = \frac{1}{M}\sum\limits_{j=1}^{M}FL(S_{B_j}, c_j).
\label{Eq:L_{bag}}
\end{aligned}
\end{equation}
\vspace{-7pt}

\textbf{Annotation Loss.} Due to the lack of explicit positive samples for supervision in MIL, the network sometimes focuses on the points outside the instance region and mistakenly regards them as the foreground.
Therefore, we introduce the annotation loss $L_{ann}$, that gives the network accurate positive samples for supervision via annotated points, to guide MIL training. $L_{ann}$ can guarantee a high score of the annotated point and mitigate mis-classification to some extent. 
Firstly, the classification score of $S_{a_j}(j \in {1, 2, ...M})$ of $a_j$ is calcluated as:
\vspace{-7pt}
\begin{equation}\small
\begin{aligned}
S_{{a_j}} &= \sigma_1(fc^{cls}(F_{a_j})) \in \mathbb{R}^{K}.
\label{Eq:S_{a_j}}
\end{aligned}
\vspace{-5pt}
\end{equation}
$L_{ann}$ is calculated with focal loss as:
\vspace{-8pt}
\begin{equation}\small
\begin{aligned}
L_{ann} &= \frac{1}{M}\sum\limits_{j=1}^{M} FL(S_{a_j}, c_j).
\label{Eq:SL_{a_j}}
\end{aligned}
\end{equation}
\vspace{-10pt}

\textbf{Negative Loss.} The conventional MIL adopts binary log loss, and it views the proposals belonging to other categories as negative samples. 
For lacking of explicit supervision from samples in background, the negative samples are not well suppressed during MIL training.
Therefore, based on $Neg_k$, the negative loss $L_{neg}$, the negative part of focal loss, is calculated as follows, where we set $\gamma=2$.
\vspace{-5pt}
\begin{equation}\small
\begin{split}
S_p &= \sigma_1(fc_{cls}(F_p)) \in \mathbb{R}^{K}; \\
L_{neg} &= \frac{1}{M}\sum\limits_{k=1}^{K} \sum\limits_{p\in Neg_{k}} c_{j,k} S_{p, k}^{\gamma} \log(1 - S_{p, k}).
\label{Eq:S_{Neg}}
\end{split}
\end{equation}
\vspace{-5pt}

\begin{figure*}[tb!]
  \hfill
  \begin{subfigure}{1.0\linewidth}
    \centering
    \includegraphics[height=0.147\linewidth]{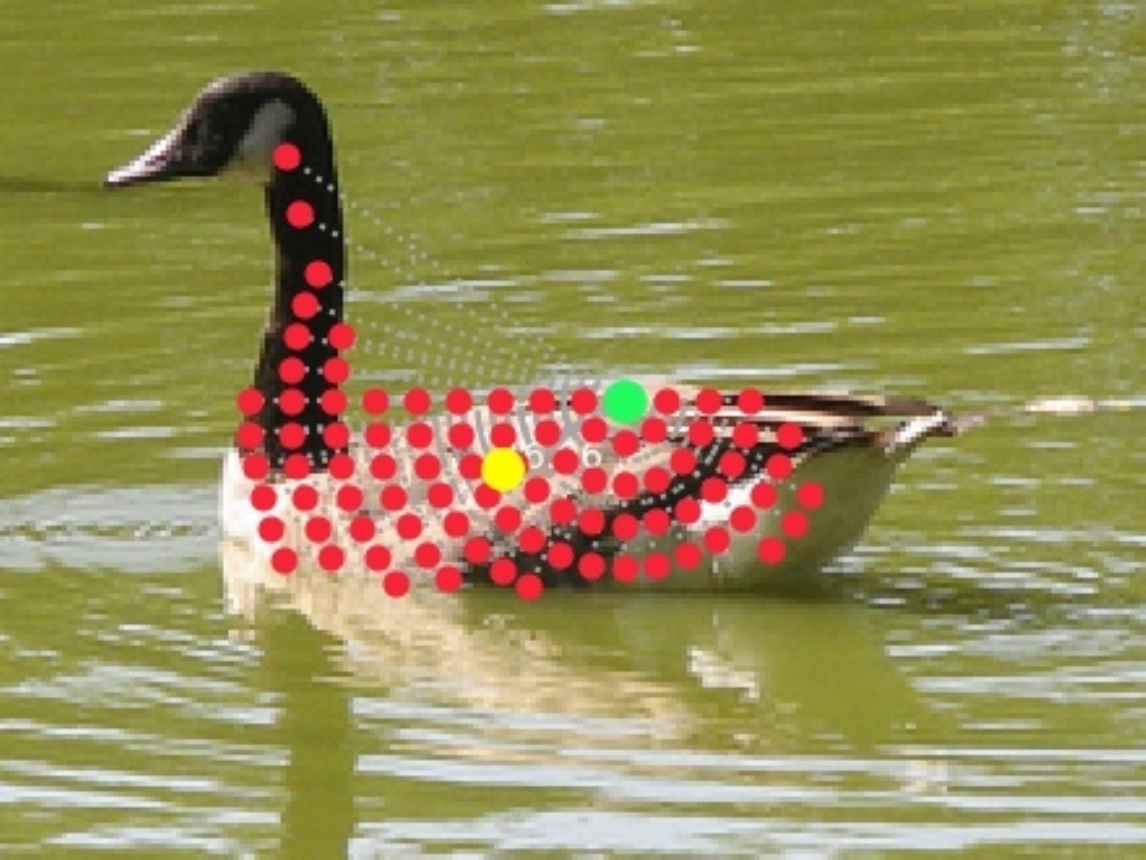}
    \includegraphics[height=0.147\linewidth]{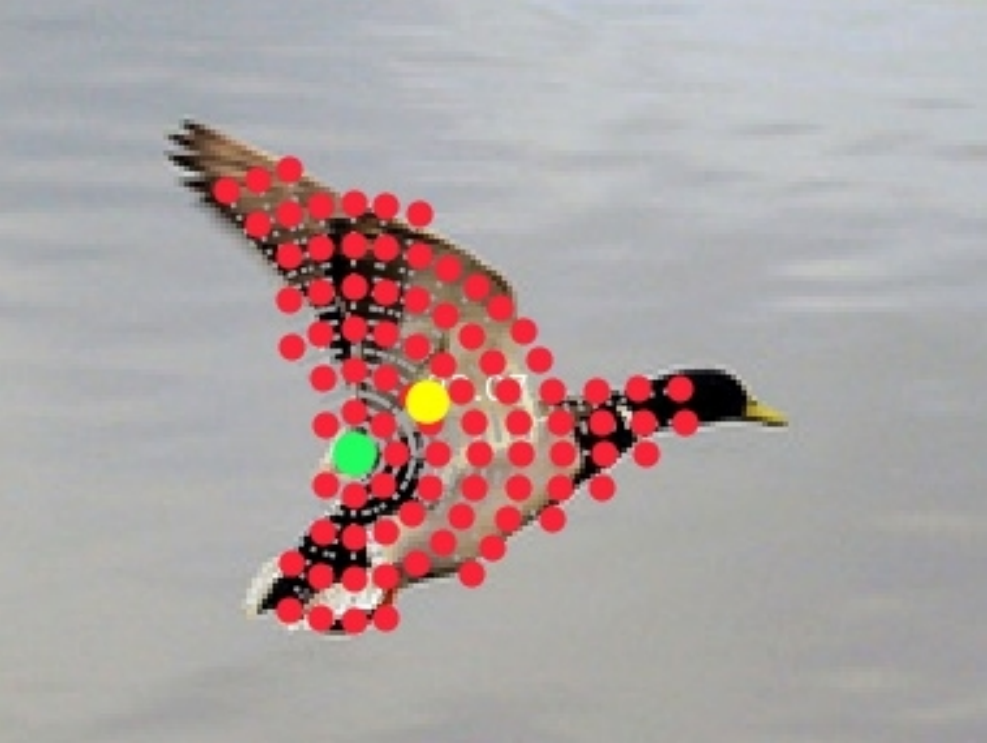}
    \includegraphics[height=0.147\linewidth]{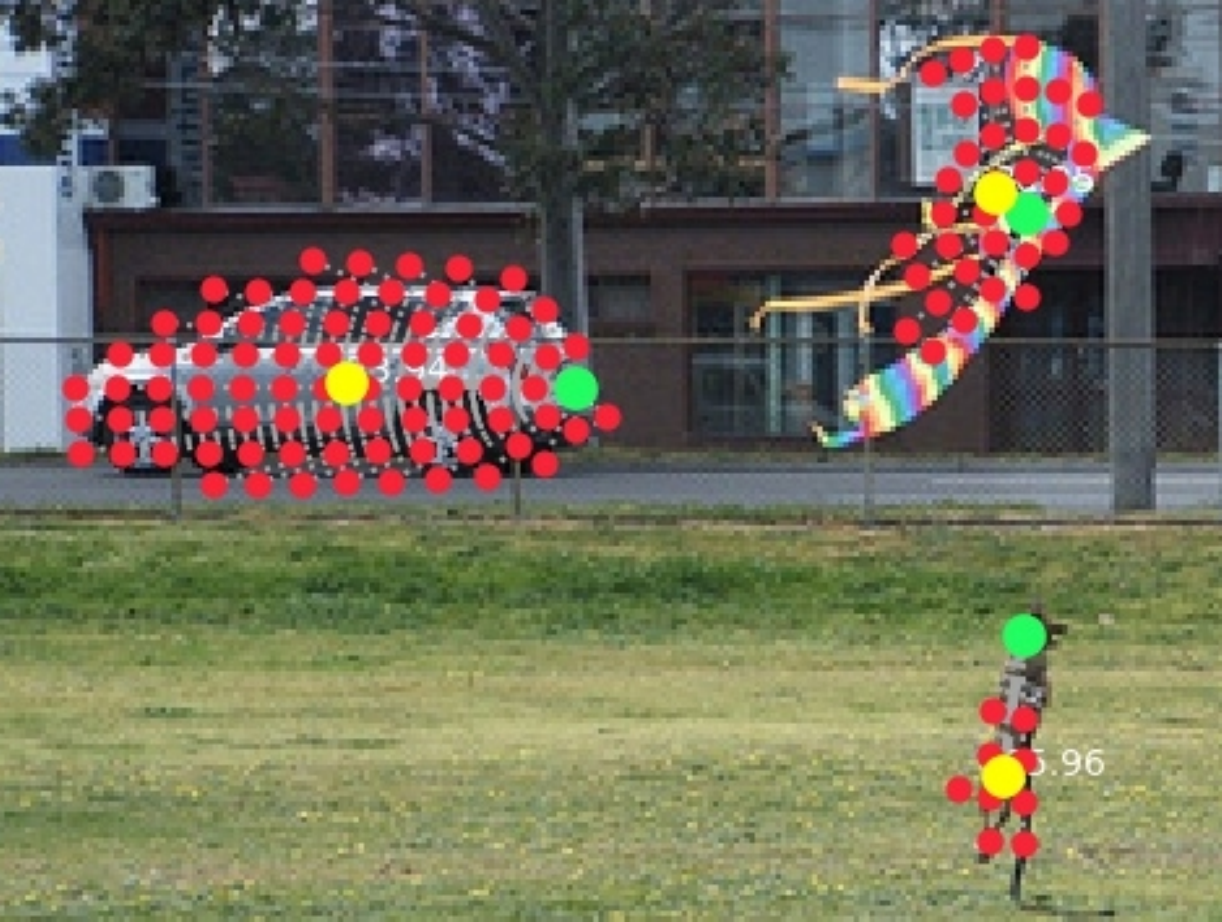}
    \includegraphics[height=0.147\linewidth]{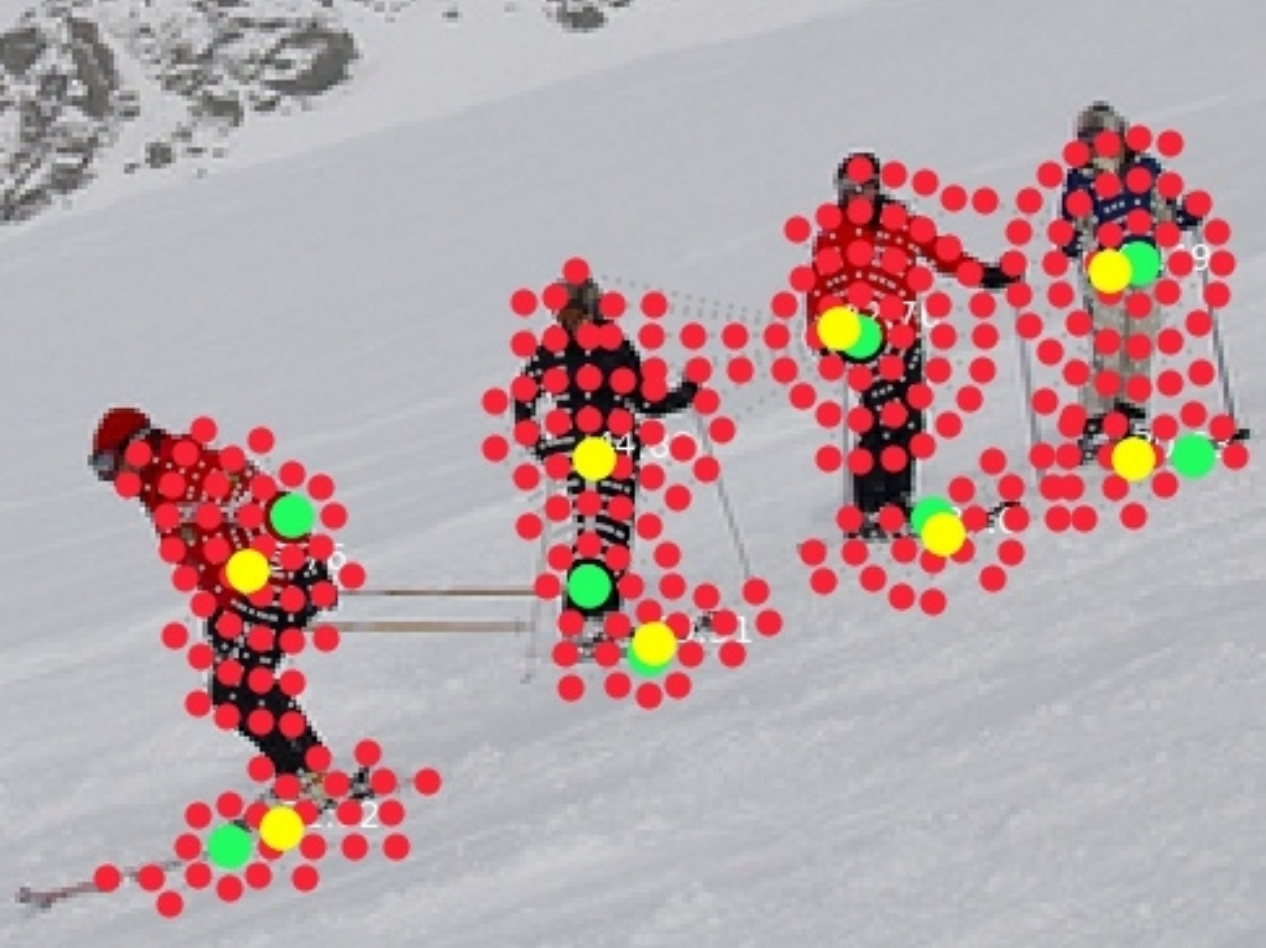}
    \includegraphics[height=0.147\linewidth]{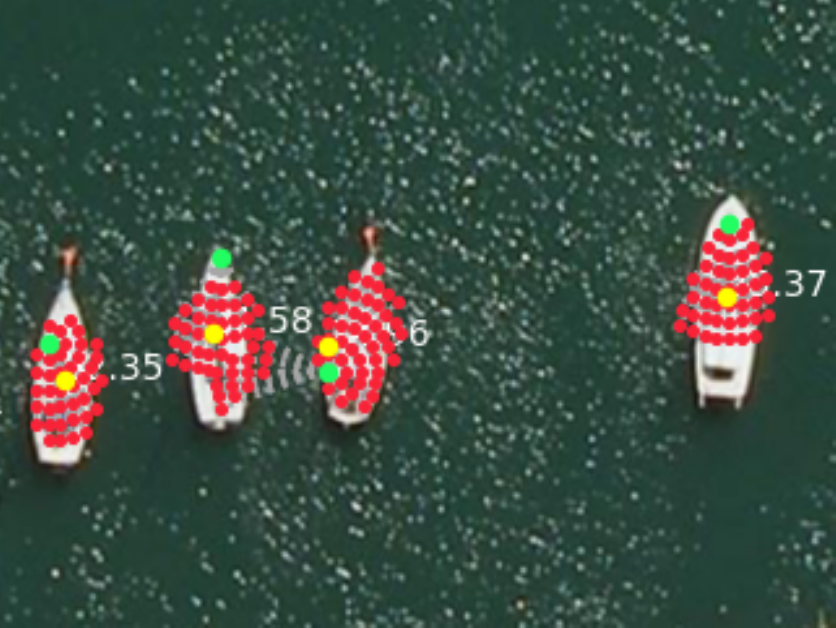}
    \label{fig:short-b}
  \end{subfigure}
  \vspace{-20pt}
  \caption{Visualization of CPR. Semantic points (red) around the annotated point (green) are weighted averaged to obtain the semantic center (yellow) as final refined point (see Sec.~\ref{sec: Refinement}). The two images on the left show the results of birds with multiple poses; the three images on the right show results of objects in multiple categories. Images are cut from the raw ones (in COCO/DOTA) for better view.}
  \vspace{-15pt}
\label{fig:visilization of CPR}
\end{figure*}

\subsection{Point Refinement} \label{sec: Refinement}
As described in Algorithm~\ref{Alg: refinement}, the trained CPRNet $\hat{E}$ is used to refine the annotated point. 
Based on $B_j$, $[S^{cls}_{B_j}]_p$ predicted by $\hat{E}$ and the constrains (details given in following), points with the same category (similar semantic) with the annotated point are selected, denoted as $B^+_j$. Then, the semantic center (final refined point), used to replace the annotated point, is set as the weighted mean of points in $B^+_j$. 

To obtain $B^+_j$, three constraints (line in blue in Algorithm~\ref{Alg: refinement}) are introduced. Constraint I is to delete points with small classification scores. We filter out the point $p \in B_j$ whose $S_{p, k_j}$ is smaller than the threshold $\delta_1$ (set as 0.1 by default) or $\delta_2*S_{a_j, k_j}$, where $\delta_2$ is set as 0.5 by default and $k_j$ is category label (not one-hot format) of $j$-th object. 
Constraint II is to delete the points that are not classified correctly. Specifically, the correct classification means the classification score $S_{p, k_j}$ of point $p$ on the given annotated category $k_j$ is higher than the scores on other categories. 
Constraint III is to delete the points closer to other object in the same category. Since two adjacent objects of the same category may interfere with each other. With the three constraints, the remaining points construct the $B^+_j$ and are weighted average to obtain the semantic center point as the final refined point, which is used as the supervision to train P2PNet~\cite{Song_2021_ICCV}.
P2PNet is the SOTA baseline for the POL task and will be specifically described in experiment section.

With the point sampling, CPRNet training and Point refinement mentioned above, the CPR can effectively mitigate semantic variance, as shown in Fig. ~\ref{fig:visilization of CPR}.
\begin{algorithm}[tb!]
{
\small
\caption{Point Refinement}
\label{Alg: refinement}
\textbf{Input:} Trained CPRNet $\hat{E}$, input image $I$, annotated points $A$, category label of annotated points $C$. \\
\textbf{Output:} Refined points $\hat{A}$.\\
\textbf{Note:} $\delta_1, \delta_2$ are thresholds. $K$ is the number of categories. $S_{p,k}$ is the predicted score on $k$-th category of point $p$. $k_j \in \{1, 2, ...K\}$ is the category label (not one-hot format) of the j-th object.\\
\vspace{-10pt}
\begin{algorithmic}[1]
\STATE $\hat{A} \leftarrow \{\}$;
\STATE $\mathbf{F} \leftarrow extract\_feature(I; \hat{E})$ according to Sec. \ref{sec:CPRNet architecture};
\FOR{$a_j \in A, c_j \in C$}
\STATE find $k_j \in \{1, 2, ..K\}$ s.t. $c_{jk_j} = 1$;
\STATE $B^+_j \leftarrow \{a_j\}$;
\STATE $S_{a_j} \leftarrow \sigma_1 (fc^{cls}(\mathbf{F}_{a_j}; \hat{E})) \in \mathbb{R}^{K}$;
\STATE $B_j \leftarrow bag\_sampling(a_j)$ according to Eq.~\ref{Eq:Bag_j};
\FOR{$p \in B_j$}
\STATE $S_p \leftarrow \sigma_1 (fc^{cls}(\mathbf{F}_p; \hat{E})) \in \mathbb{R}^{K}$;
\STATE $s_p \leftarrow S_{p,k_j}$;
\IF{\textcolor{blue}{$s_p > \delta_1$ and $s_p > \delta_2 * S_{a_j, k_j}$ and
\\ \ \ \ \ $k_j = argmax_{1 \leq k\leq K}\ S_{p, k}$ and
\\ \ \ \ \ $a_j = argmin_{a \in A}\ ||p-a||$\ \ \ }
} 

\STATE $B^+_j \leftarrow B^+_j \cup \{ p \}$;
\ENDIF
\ENDFOR
\STATE $\hat{a_j} \leftarrow (\sum_{p\in B^+_j} S_p * p) / (\sum_{p\in B^+_j} S_p)$;
\STATE $\hat{A} \leftarrow \hat{A} \cup \{\hat{a_j}\}$;
\ENDFOR
\end{algorithmic}
}
\end{algorithm}
\vspace{-12pt}

\section{Experiment}

\subsection{Experimental Settings}
\textbf{Datasets.}
For experimental comparisons, three public available datasets are used for point supervised localization task: COCO~\cite{lin2014microsoft}, DOTA-v1.0~\cite{DBLP:conf/cvpr/XiaBDZBLDPZ18} and SeaPerson. 
\textbf{COCO} is MSCOCO 2017, and it has 118k training and 5k validation images with 80 common categories. Since the ground-truth on the test set are not released, we train our model on the training set and evaluate it on the validation set. 
\textbf{DOTA}(v1.0) provides 2,806 images with 15 object categories. We utilize training set for the training and validation set for evaluation.
\textbf{SeaPerson}\footnote{SeaPerson is a low-resolution tiny person dataset and disclose little personal privacy from the appearance.} is a dataset for tiny person detection collected through a UAV camera at the seaside. The dataset contains 12,032 images and 619,627 annotated persons with low resolution. The images in the SeaPerson are randomly selected as training, validation, and test sets with the proportion of 10:1:10. The details are given in the supplemental materials.

\textbf{Coarse Point Annotation} \label{Sec: Coarse Point Annotation}
\label{sec3}
In practical scenarios, the coarse point can be obtained through annotating any single point on an object. However, since datasets in the experiment are already annotated with masks or bounding boxes, according to the law of large numbers, it is reasonable that the manually annotated points follow Gaussian distribution. Furthermore, since the annotated points must be inside the bounding box or mask of the object, 
then an improved Gaussian distribution, named as Rectified Gaussian (RG) Distribution, is utilized for annotation. $RG(p;0, \frac{1}{4})$ is chosen to generate the point annotations for the experiments.

\vspace{-5pt}
\begin{equation}\small
\begin{aligned}
& \phi(p; \mu, \sigma) = Gauss(p;\mu, \sigma) \cdot Mask(p); \\
& RG(p; \mu, \sigma) = \frac{\phi(p;\mu, \sigma)} {\int_{p} \phi(p; \mu, \sigma)}.
\label{Eq:point generation}
\end{aligned}
\end{equation}
\vspace{-2pt}
where $\mu$ and $\sigma$ are the mean and standard deviation of Gaussian distribution, respectively. $Mask(p)\in \{0, 1\}$ denotes whether point $p$ falls inside the mask of an object. If it is generated from the bounding box annotation, then the box is treated as a mask.

\textbf{Evaluation.} Similar to WSOD, a point-box distance, calculated between the point and box, is used for evaluation. Specifically, the distance $d$ between point $a=(x, y)$ and  bounding box $b=(x^c, y^c, w, h)$ is defined as:
\begin{equation}\small
d(a, b)={\sqrt{\left(\frac{x-x^c}{w}\right)^2+\left(\frac{y-y^c}{h}\right)^2}}.
\label{Eq:distance-base}
\end{equation}
where $(x^c, y^c)$, $w$, $h$ are the center point, width, and height of the bounding box, respectively. 
The distance $d$ is used as the matching criterion for POL performance. 
A point and the object's bounding box are matched if the distance $d$ is smaller than a predefined threshold $\tau$. ($e.g.$ $\tau=1.0$ means that as long as the point falls within an matched ground-truth box, the point successfully matches the ground-truth box.) If a bounding box has multiple matched points, the point with the highest score is chosen. While a point has multiple matched objects, the object with the smallest point-box distance is selected. A true positive (TP) is counted if a point matches an object. Otherwise, a false positive (FP) is counted. Neither TP nor FP will be counted if a point matches an object that is annotated as ignore, which follows the evaluation criteria of pedestrian detection~\cite{zhang2017citypersons} and TinyPerson benchmark~\cite{Yu2020ScaleMF}. We adopt mean Average Precision with $\tau=1.0$ ($mAP^{all}_{1.0}$) as the main metric for experimental comparisons. We do not consider a small $\tau$ because it makes the task more like center localization instead of object localization. Here we also report the results of $\tau=0.5$ and $\tau=2.0$ in Table~\ref{Eq:CPR basic loss}, which can be informative.

\textbf{Implementation Details for CPRNet.} Our codes are based on MMDetection~\cite{mmdetection}. Same to the default setting of the object detection on COCO, the stochastic gradient descent (SGD ~\cite{DBLP:series/lncs/Bottou12})
algorithm is used to optimize in 1x training schedule. The learning rate is set to 0.0025 and decays by 0.1 at the 8-th and 11-th epoch, respectively. 

\subsection{Experimental Comparisons}
The POL task with coarse point annotation is divided into two key parts: refining the coarse point annotation and training the point localizer with refined points. 

\textbf{Detector with Pseudo Box.} \label{sec:Detector with Pseudo Box} For training a point localizer, an intuitive idea is to convert the point-to-point (POL) to a box-to-box (object detection) problem. Firstly, a fixed-size pseudo-box is generated with each annotated point as the center. Next, the pseudo-box is used to train a detector. Finally, during inference, the center points of the boxes predicted by the trained detector are used as the final output. Following ~\cite{DBLP:conf/cvpr/RiberaGCD19}, we conduct the pseudo box for localization and give the performance in row 1 of Table~\ref{tab:main}. The difference to~\cite{DBLP:conf/cvpr/RiberaGCD19} is that the RepPoint~\cite{DBLP:conf/iccv/YangLHWL19} is utilized rather than Faster RCNN~\cite{DBLP:conf/nips/RenHGS15}, due to its efficiency.

\textbf{Multi-Class P2PNet.} \label{sec:p2p}
We adopt P2PNet\footnote{In our experiments, we re-implement P2PNet and further endow it the new ability of handling multi-class prediction, which to our best efforts, aligns the results with those reported in the raw paper~\cite{Song_2021_ICCV}.}, to train with point annotation and predict point for each object during inference, as a
stronger baseline for POL task. Improvement can be made, especially when there are multiple categories:
\romannumeral1) The backbone of P2PNet in  this paper is Resnet-50 rather than VGG16~\cite{DBLP:journals/corr/vgg}. 
\romannumeral2) Instead of using the Cross-Entropy loss, we adopt the focal loss when optimizing classification to better deal with the problem of imbalance; 
\romannumeral3) The Smooth-$\ell_1$ loss instead of $\ell_2$ loss is used for regression. 
\romannumeral4) In label assignment, different from a one-to-one matching in the default P2PNet, we assign top-k positive samples for each ground-truth and regard remaining samples as background. And then the NMS ~\cite{DBLP:conf/icpr/NeubeckG06} post-processing for points is performed to obtain the final point results.
The performances of P2PNet, given in row 2 in Table~\ref{tab:main}, improves a lot compared with the pseudo box (row 1 in Table~\ref{tab:main}). P2PNet is a stronger baseline for POL task.

\begin{figure}[tb!]
\begin{center}
    \begin{tabular}{ccc}
    \includegraphics[width=0.98\linewidth]{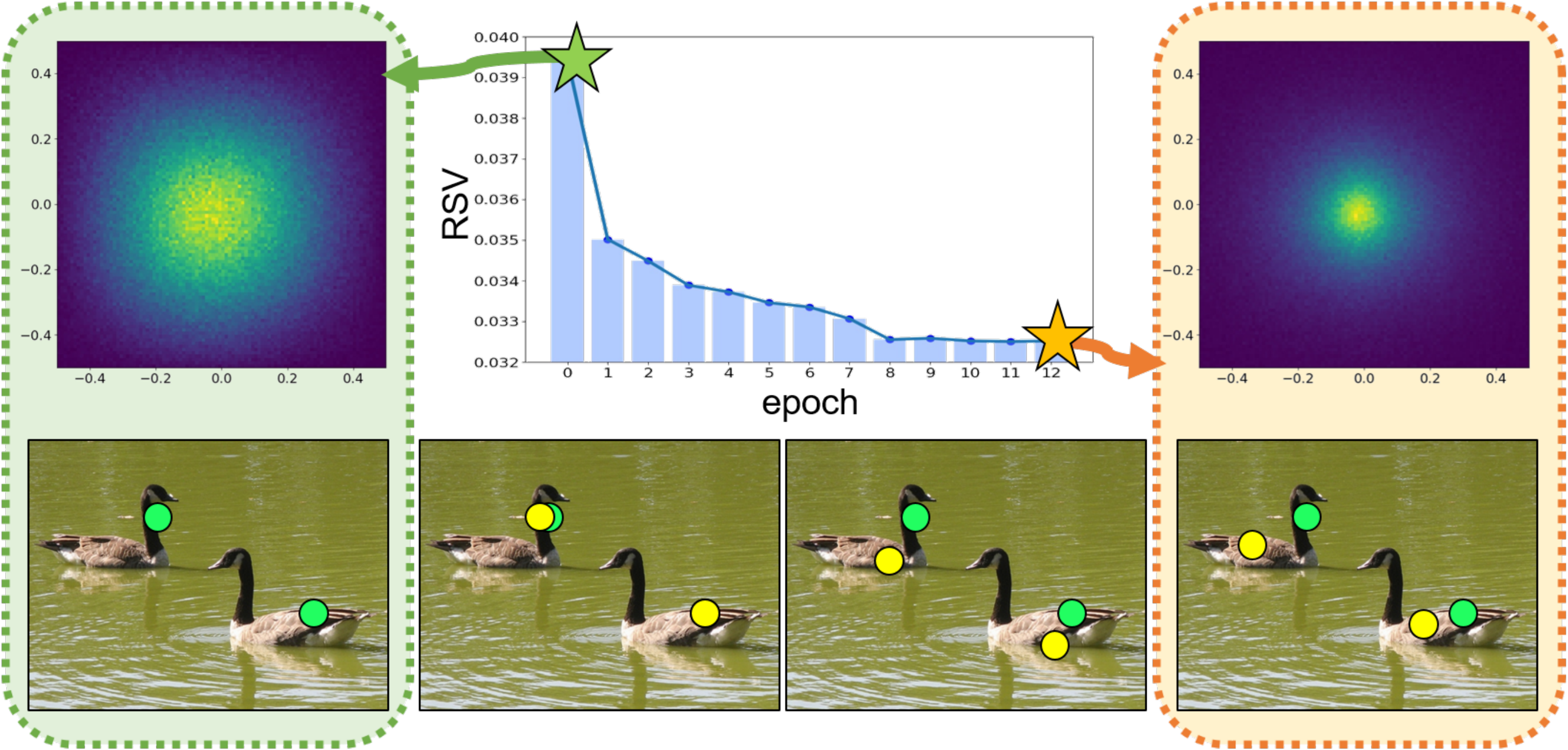}
    \end{tabular}
\vspace{-12pt}
\setlength{\belowcaptionskip}{-0.2cm}
\caption{The top-left and top-right figures are the relative position distribution (Eq.~\ref{Eq:relative position distribution}) heatmap of annotated points and final refined points in bounding boxes. The top-middle figure is RSV curve of refined points (Eq.~\ref{Eq:a semantic variance})) in each epoch during CPRNet training. The bottom four figures give the annotated points (green) and refined points (yellow) in four epochs, showing the refined points gradually converge to the semantic center points.}
\label{fig:alliave smentic variance}
\end{center}
\vspace{-19pt}
\end{figure}

\begin{table}[tb!]
\begin{center}
    \resizebox{0.48\textwidth}{!}{
\begin{tabular}{c|c|c|c|c}
\hline
refinement & localizer & COCO & DOTA& SeaPerson                \\
\hline
\hline
- & RepPoint$^*$  & 37.42 &   47.22   & 47.72 \\
- & P2PNet       & 38.48 & 48.34 & 76.52 \\
\hline
Self-Refinement (ours) & P2PNet   & 50.86  & 60.39 & 84.96  \\
CPR (ours)             & P2PNet  & \textbf{55.46} & \textbf{63.81} & \textbf{85.86} \\
\hline
\end{tabular}}
\end{center}
\vspace{-15pt}
\caption{The experimental comparisons ($mAP^{all}_{1.0}$) of localizers in three datasets: COCO, DOTA and SeaPerson. RepPoint$^*$ means RepPoint with pseudo box (details in Sec.~\ref{sec:Detector with Pseudo Box}).}
\vspace{-5pt}
\label{tab:main}
\end{table}

\begin{table}[tb!]
\begin{center}
\resizebox{0.4\textwidth}{!}{
\begin{tabular}{l|c|c|c|c|c|c}
\hline
pos & MIL & ann & neg & $mAP^{all}_{1.0}$ & $mAP^{all}_{0.5}$ & $mAP^{all}_{2.0}$ \\
\hline
\hline
& \checkmark &            &            & 39.07 & 28.37 & 46.90 \\
& \checkmark & \checkmark &            & 39.45 & 28.22 & 47.42 \\
& \checkmark &            & \checkmark & 54.24 & 46.56 & 58.94 \\
&           & \checkmark & \checkmark  & 51.82 & 46.24 & 55.67 \\
\checkmark& & \checkmark & \checkmark  & 42.72 & 33.07 & 48.81 \\
& \checkmark & \checkmark & \checkmark & \textbf{55.46} & \textbf{50.23} & \textbf{59.49} \\
\hline
\end{tabular}
}
\end{center}
\vspace{-15pt}
\setlength{\belowcaptionskip}{-0.5cm}
\caption{The effect of training loss in CPRNet: MIL loss, annotation loss, negative loss. The pos loss is for comparison.}\vspace{-5pt}
\label{tab:training loss}
\end{table}

\textbf{Self-Refinement.} For refining the coarse point annotation, inspired by ~\cite{DBLP:journals/corr/abs-1805-02641}, we propose a self-refinement technique that works as a strategy based on self-iterative learning. Firstly, the aforementioned pseudo box strategy is adopted to train a point localizer. Then, the weighted mean of the points predicted by the localizer works as the new supervision. Finally, the refined points are obtained. With these refined points as supervision, the performance of P2PNet as point localizer is given in row 3 in Table~\ref{tab:main}, where it alleviates the semantic variance problem.

\begin{table}[tb!]
\begin{center}
\resizebox{0.35\textwidth}{!}{
    \begin{tabular}{l|c|c|c}
    \hline
    detector & COCO & DOTA & SeaPerson \\
    \hline
    \hline
    RetinaNet        & 32.61 & 51.53 & 48.50 \\
    FasterRCNN       & 35.29 & 51.15 & 47.93 \\
    RepPoint         & 37.42 & 47.26 & 47.72 \\
    \hline
    RetinaNet w. CPR & 51.35 & 63.69 & 77.90 \\
    FasterRCNN w. CPR& 53.21 & 63.00 & 77.80 \\
    RepPoint w. CPR  & 53.97 & 60.37 & 78.94 \\
    \hline
    \end{tabular}
}
\end{center}
\vspace{-10pt}
\caption{$mAP^{all}_{1.0}$ of more architectures on the three datasets.}
\label{tab:arc}
\end{table}

\textbf{CPR.} Compared with self-refinement, CPR (shown in  row 5 of Table~\ref{tab:main}) obtains more performance gain, indicating it is more efficient for dealing with the semantic variance. To quantify the semantic variance of point annotation, the relative semantic variance (RSV), which is calculated based on the relative distance of the point to the center point: 
\vspace{-8pt}
\begin{equation} \small
\begin{aligned}
x' =& \frac{x - x^c}{w};  \quad
y' = \frac{y - y^c}{h}; \\
RSV &= [Var(x') *  Var(y')]^{\frac{1}{2}}.\\
\label{Eq:a semantic variance}
\end{aligned}
\vspace{-18pt}
\end{equation}
\vspace{-10pt}

\noindent where $(x, y)$ is an annotated point or refined point and $(x^c, y^c)$ is the corresponding center point in the bounding box of an object. $Var(x')$ and $Var(y')$ are the variance of $x'$ and $y'$ of all objects in the dataset, respectively. 
Statistically, the smaller RSV means the $(x, y)$ holds a more stable relative position to its corresponding $(x^c, y^c)$, as shown in Eq.~\ref{Eq:a semantic variance}. Considering the $(x^c, y^c)$ as a strict key point, the intuition behind the RSV is that a small RSV for a category is equivalent to a strict annotation, which can effectively reduce the semantic variance of annotations. As shown in Fig.~\ref{fig:alliave smentic variance}, the annotated coarse point holds a larger RSV, while the refined point via CPR obtains a smaller RSV.

To show the relative position distribution of annotated points in the bounding boxes, we calculate $Prob(x', y')$ as:
\vspace{-5pt}
\begin{equation}\small
\begin{aligned}
Prob(x'=p_x, y'=p_y) = \frac{\sum\limits_{1\leq j\leq M} \mathbf{\mathbb{I}}\{x'_j=p_x'\ and \ y'_j=p_y'\}}{M}.
\label{Eq:relative position distribution}
\end{aligned}
\end{equation}
where $(x'_j, y'_j)$ is relative position of annotated point or refined point for object $j$ in dataset, $\mathbf{\mathbb{I}}\{*\}$ is 1 if $*$ is true, otherwise 0. $Prob(x', y')$ is shown as the heatmap in Fig.~\ref{fig:alliave smentic variance}.

\begin{table*}[tb!]
\begin{center}
    \begin{subtable}[tb]{0.18\linewidth}
        \resizebox{1.0\textwidth}{!}{
        \begin{tabular}{l|c|c}
        \hline
        feature & $R$ &$mAP^{all}_{1.0}$ \\
        \hline
        \hline
        & 5  & 53.32 \\
        & 8  & \textbf{55.46} \\ 
        P3 & 10 & 55.19 \\
        & 15 & 55.38 \\
        & 20 & 55.04 \\
        & 25 & 53.85 \\
        \hline
        \end{tabular}
        }
        \caption{Different $R$ with P3}
        \label{tab:sample_r P3}
    \end{subtable}
    \begin{subtable}[tb]{0.18\linewidth}
        \resizebox{1.0\textwidth}{!}{
        \begin{tabular}{l|c|c}
        \hline
        feature & $R$ &$mAP^{all}_{1.0}$ \\
        \hline
        \hline
        & 5 & 48.64 \\
        & 10  & 53.76 \\
        P2 & 15 & 54.26  \\
        & 20 & \textbf{54.64} \\
        & 30 & 54.24 \\
        & 40 & 53.11 \\
        \hline
        \end{tabular}
        }
        \caption{Different $R$ with P2}
        \label{tab:sample_r P2}
    \end{subtable}
    \hspace{2em}
    \begin{subtable}[tb]{0.24\linewidth}
    \resizebox{0.95\textwidth}{!}{
        \begin{tabular}{l|c|c|c|c}
        \hline
        \multicolumn{2}{c|}{\bf I} & \multirow{2}{*}{\bf II} & \multirow{2}{*}{\bf III} & \multirow{2}{*}{$mAP^{all}_{1.0}$} \\
        \cline{1-2}
        $\delta_1$ & $\delta_2$ &  &  &  \\
        \hline
        \hline
        0 & 0.5 & \checkmark & \checkmark & 45.17 \\
        0.1  & 0 & \checkmark & \checkmark & 54.96 \\
        0.1  & 0.5 &  & \checkmark & 54.25 \\
        0.1 & 0.5 & \checkmark &  & 52.69 \\
        0.1  & 0.5 & \checkmark & \checkmark & \textbf{55.46} \\
        \hline
        \end{tabular}
    }
    \caption{Constraints in refinement.}
    \label{tab:refinement}
    \end{subtable}
    \hspace{1em}
    \begin{subtable}[h]{0.24\linewidth}
    \centering
    \begin{subtable}[t]{0.8\linewidth}
    \centering
    \resizebox{1.0\textwidth}{!}{
        \begin{tabular}{l|c|c}
        \hline
        annotation & CPR & $mAP^{all}_{1.0}$ \\
        \hline
        \hline
        coarse &  & 38.48\\
        coarse & \checkmark &55.46 \\
        center & & 57.47 \\
        \hline
        \end{tabular}
    }
    \vspace{0pt}
    \caption{Different annotation.}
    \vspace{0pt}
    \label{tab:compare with center annotation.}
    \end{subtable}
    \begin{subtable}[t]{1.0\linewidth}
        \resizebox{1.0\textwidth}{!}{
        \begin{tabular}{l|c|c}
        \hline
        CPR & P2PNet & $mAP^{all}_{1.0}$ \\
        \hline
        \hline
        ResNet-50  & ResNet-50  & 55.46 \\
        ResNet-50  & ResNet-101 & 55.80 \\
        ResNet-101 & ResNet-101 & 56.43 \\
        \hline
        \end{tabular}
        }
    \vspace{0pt}
    \caption{Different backbones.}
    \vspace{0pt}
    \label{tab:general}
    \end{subtable}
    \end{subtable}
\end{center}
\vspace{-25pt}
\caption{Ablation studies.}
\vspace{-20pt}
\end{table*}

\textbf{Performance Analysis.}
Pseudo box based localizer is almost equivalent to train a detector that treats the points in the neighborhood of the annotated points as positive samples and others as negative samples. The general detectors perform label assignment with IoU, which depends heavily on the scale information of the given bounding box. However, precise bounding box can not be obtained from point annotation of POL, leading to poor performance of Pseudo box based localizer. 
P2PNet adopts hungarian algorithm to achieve a purely point-to-point assignment, obtaining better performance than pseudo box based localizer. 
However, P2PNet is much sensitive to the accuracy of annotated points and the semantic variance. 
Therefore, point refinement strategy, effectively reducing the semantic variance of the annotation, achieves better performance. CPR that can better capture the semantic information, outperforms. 

\vspace{-2pt}
\subsection{Ablation Studies}
\vspace{-2pt}
To further analyze CPR's effectiveness and robustness, we conduct more experiments on COCO.

\textbf{Training Loss in CPRNet.}  
Ablation study of the training loss is given in Table~\ref{tab:training loss}. 
The CPR loss given in row 6 in~Table~\ref{tab:training loss} obtains 55.46 mAP. \textbf{i) MIL loss.} If the MIL loss is removed (row 4), the CPRNet training relies on the annotation loss and the negative loss, the performance drops 3.64 points (51.82 \emph{vs} 55.46). When we replace the MIL loss with the pos loss, which treats all the sampled points in the MIL bag as positive samples (line 5), the performance sharply declines by 12.74 points (42.72 \emph{vs} 55.46)
, showing that MIL can autonomously discern points belonging to the object. 
\textbf{ii) Annotation loss.} Lacking of the annotation loss (row 3), the performance of localization decreases 1.22 points (54.24 \emph{vs} 55.46). The annotation loss guides the training through an given accurate positive supervision.
\textbf{iii) Negative loss.} 
With the negative loss (row 2), the performance improves by 16.01 points (55.46 \emph{vs} 39.45), indicating that only MIL loss is not enough to suppress the background, and the negative loss is inevitable.

\textbf{Feature Map Level.} 
The CPRNet is established based on single level feature map of FPN. Table~\ref{tab:sample_r P3} and~\ref{tab:sample_r P2} show the performance with different feature map levels. Since the performance on P3 is similar to that of P2, P3 is chosen for our experiments in COCO if not otherwise specified.

\textbf{Sampling Scope.}
Table~\ref{tab:sample_r P3} and~\ref{tab:sample_r P2} show the performance of different radius $R$, where $R$ is a sensitive hyper-parameter in CPRNet.  On P3, the best performance 55.46 is obtained when $R$ is set as 8. If the sampling scope reduces, such as $R=$ 5, the performance significantly declines to 53.32, since the sampling scope is limited to a small local region, leading to a worse refinement. While the scope getting larger, the performance becomes steady but drops slowly until $R$ is over 25 (53.85), since the bag $B_j$ for MIL introduces more noise, which degrades the performance.


\textbf{Point Refinement Policy.} For point refinement, there are three constraints (described in Sec.~\ref{sec: Refinement}).
$\delta_1$ and $\delta_2$ are threshold of constraint I. 
In Table~\ref{tab:refinement}, it shows that the three constraints together obtain performance gain.

\textbf{Upper Bound Analysis.} To further validate the CPR, a comparison between CPR and a strict annotation based localizer, which can be seen as the upper bound for CPR, is conducted on COCO. 
Since it is hard to annotate the objects in general dataset (e.g. COCO) with key points. Therefore, we approximately use the center point of each objects' bounding box as a kind of strict point annotation. The experiment results in Tabel~\ref{tab:compare with center annotation.} show that CPR can achieve a comparable performance to center point annotation based localizer (55.46 \emph{vs} 57.47).

\textbf{Localizer Architecture.} Table~\ref{tab:arc} shows that CPR can further improve the performance of different localizers, such as Faster-RCNN, RetinaNet and RepPoint.

\textbf{Backbone}. As shown in Table~\ref{tab:general}, due to the stronger backbone Resnet-101 for CPRNet and P2PNet, it obtains a better performance 56.43.

\vspace{-8pt}
\section{Conclusion and Outlook}
\vspace{-4pt}
In this paper, we rethink the semantic variance problem in point-based annotation caused by the non-uniqueness of optional annotated points. The proposed CPR samples points in neighbourhood, finds the semantic points on the object by introducing MIL, 
and then weighted averages these semantic points to obtain the semantic center of the object as the supervision for the localizer.
CPR alleviates semantic variance and facilitates the extension of POL task to multi-class and multi-scale. Comprehensive ablations on multiple datasets further verify the effectiveness of our model. In future, we will study on an adaptive $R$ and explore the possibilities of extending CPR to other tasks.

\textbf{Limitation.}
The performance is sensitive to $R$, which is not an adaptive value in this paper and may limit CPR to better deal with the multi-scale of objects to some extent.

\textbf{Broader Impact.} Similar to most of object detection and localization task, the bias of dataset from the intrinsic artifacts are not considered.

\vspace{-12pt}
\small\section{Acknowledgements}
\vspace{-5pt}
This work was supported in part by the Youth Innovation Promotion Association CAS, the National Natural Science Foundation of China (NSFC) under Grant No. 61836012 and 61771447, and the Strategic Priority Research Program of the Chinese Academy of Sciences under Grant No.XDA27000000. The work was partially done during Xuehui's internship at JD Explore Academy, China. We would like to thank Jing Zhang for helpful discussion and suggestions.

{\small
\bibliographystyle{ieee_fullname}
\bibliography{paper_with_appendix}
}

\clearpage

\begin{appendices}

\begin{center}{\bf \Large Appendix}\end{center}\vspace{-2mm}

\section{More experiment.}

\textbf{Sample shape and sample density.} In Table~\ref{tab: sample denstity}, we ablate on sample density $u_0$ and validate that $u_0$ has little impact. 
Theoretically, circles are orientation equivalent, which makes it suitable to model multi-view object. Experimentally, the sampling results of circle and rectangle are very close due to the discrete sampling, leading to comparable performance (Table~\ref{tab: sample selection shape}, where different aspect ratio $rect._{w:h}$ are studied). 

\textbf{Threshold of refinement.} $\delta_{1}$ and $\delta_{2}$ are thresholds, well defined in detection and localization tasks. 
$\delta_{1}$, $\delta_{2}$, $u_0$ have little impact to performance (Table~\ref{tab:abl}). 

\begin{table}
\centering
    \begin{subtable}[t]{0.45\linewidth}
        \begin{center}
            \begin{tabular}{c|c}
            \hline
            $region$ & $mAP^{1.0}$ \\
            \hline
            \hline
            $circle$      & 55.46 \\
            $rect._{1:1}$ & 55.39 \\
            $rect._{2:1}$ & 54.77 \\
            $rect._{1:2}$ & 54.96 \\
            \hline
            \end{tabular}
        \caption{$region$}
        \label{tab: sample selection shape}
        \end{center}
    \end{subtable}
    \begin{subtable}[t]{0.45\linewidth}
        \begin{center}
            \begin{tabular}{c|c}
            \hline
            $u_{0}$ & $mAP^{all}_{1.0}$               \\
            \hline
            \hline
            12 & 55.25 \\
            8  & 55.46 \\
            6  & 55.36 \\
            4  & 55.42 \\
            \hline
            \end{tabular}
        \end{center}
        \vspace{-10pt}
        \caption{$u_0$}
        \label{tab: sample denstity}
    \end{subtable}
    \begin{subtable}[t]{0.45\linewidth}
        \begin{center}
        \begin{tabular}{l|c}
        \hline
        $\delta_1$ & $mAP^{all}_{1.0}$ \\
        \hline
        \hline
        0.10 & 55.46 \\
        0.15 & 55.98 \\
        0.20 & 56.04 \\
        0.25 & 55.78 \\
        \hline
        \end{tabular}
        \end{center}
        \vspace{-10pt}
        \caption{$\delta_1$}
        \label{delta1}
    \end{subtable}
    \begin{subtable}[t]{0.45\linewidth}
        \begin{center}
        \begin{tabular}{l|c}
        \hline
        $\delta_2$ & $mAP^{all}_{1.0}$ \\
        \hline
        \hline
        0.25 & 55.19 \\
        0.50 & 55.46 \\
        0.75 & 55.22 \\
        1.0  & 55.41 \\
        \hline
        \end{tabular}
        \end{center}
        \vspace{-10pt}
        \caption{$\delta_2$}
        \label{delta2}
    \end{subtable}
    \vspace{-5pt}
    \caption{Ablation study for hyper-parameters.}
    \label{tab:abl}
\end{table}

\section{SeaPerson}

SeaPerson is building for tiny person localization, which can help maritime quick rescue, beach safety inspection and so on.
The resolution of images are mainly 1920 $\times$ 1080 and the person size is extremely low (about 22.6 pixels). Therefore, there is no privacy sensitive information.

\textbf{Dataset Collection.} 
SeaPerson is collected as :
\romannumeral1) Videos are recorded in various seaside scenes by a RGB camera on a Unmanned Aerial Vehicle.
\romannumeral2) We sample an image of every 50 frames from video and remove images with high homogeneity.
\romannumeral3) We annotated all persons in all sampled images with bounding boxes.
\romannumeral4) Following the rules of coarse point annotation in Sec~\ref{Sec: Coarse Point Annotation}, coarse point annotation is obtained on SeaPerson for POL task.

\textbf{Dataset Splitting.} We randomly split dataset into three subsets (training set, valid set and test set), while images from the same video sequence cannot be separated into different subsets. 
As shown in Table~\ref{tab: volumn of datasets}, the ratio of images' number in training set, valid set and test set is about 10:1:10. 

\begin{figure}[tb!]
\begin{center}
    \begin{tabular}{lcc}
     \includegraphics[width=1.02\linewidth]{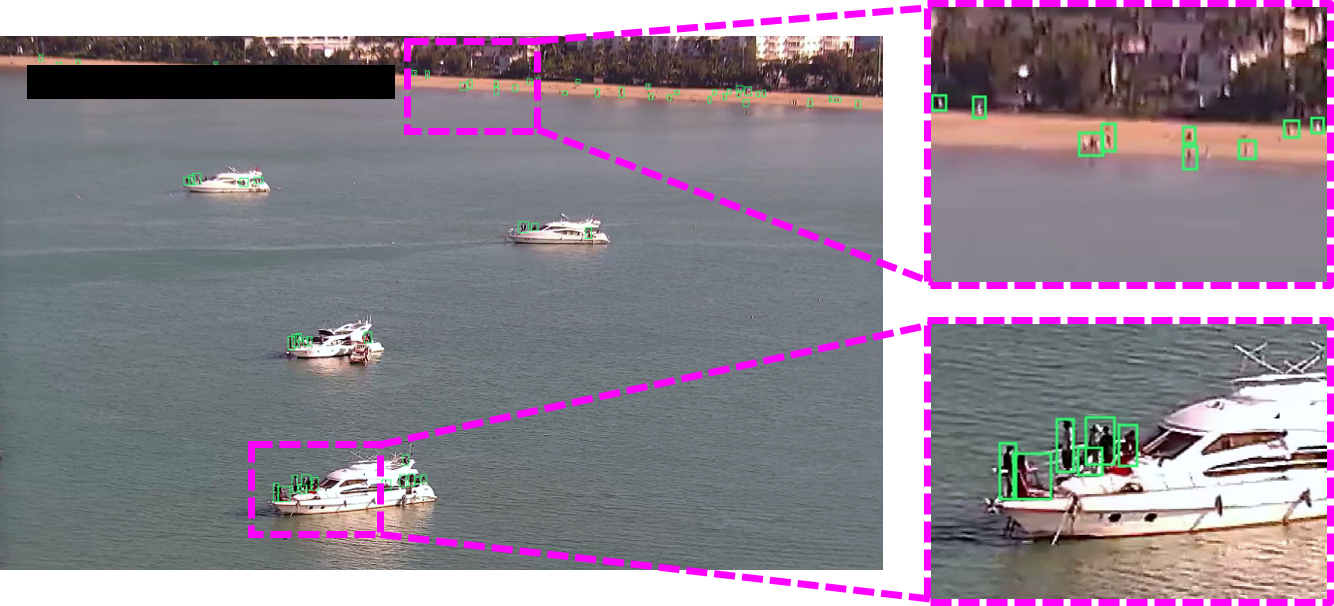}
    \end{tabular}
\vspace{-15pt}
\setlength{\belowcaptionskip}{-0.5cm}
\caption{Examples of SeaPerson.}
\label{fig: example of seaperson}
\end{center}
\end{figure}

\textbf{Dataset Properties.} Our proposed SeaPerson is similar with TinyPerson while the volumn of SeaPerson is about 7 times that of TinyPerson. The absolute size and relative size of objects are very small as shown in Table~\ref{tab: size of datasets} and Fig.~\ref{fig: example of seaperson}. In such scenario, we only care about the position of the object rather than the size of the object, which makes it very suitable for POL task. In addition, SeaPerson can also be used as a dataset for tiny object detection due to the bounding box annotation.


\section{Implementation Details for CPRNet}
ResNet-50 is used as the backbone network unless otherwise specified and FPN is adopted for feature fusion.
P2 (stride is 4) is used for SeaPerson and P3 (stride is 8) is used for COCO and DOTA. 
The mini-batch is 64/4/4 images and all models are trained with 8/2/4 GPUs for COCO/DOTA/SeaPerson. The training epoch numbers are set as 12/12/6, the learning rate are set as 0.0025, 0.00025, 0.0125 and decays by 0.1 at the 8-th/8-th/4-th and 11-th/11-th/5-th epoch for COCO/DOTA/SeaPerson, respectively. In default settings, the backbone is initialized with the pre-trained weights on ImageNet and other newly added layers are initialized with Xavier. 
The sampling radius $R$ is set as 8/7/5 for COCO/DOTA/SeaPerson by defalut.

In dataset pre-processing, for COCO dataset, the short side of the images is resized to 400, and the ratio of width and height is kept. In dota dataset, images are split into sub-images (1024 $\times$ 1024 pixels) with overlap (200 $\times$ 200 pixels). And in SeaPerson, images are split into sub-images (640 $\times$ 640 pixels) with overlap (100 $\times$ 100 pixels). For data augmentation, only random horizontal is utilized in our CPRNet training.

\begin{table}
\begin{center}
\resizebox{0.48\textwidth}{!}{
\begin{tabular}{l|c|c|c|c}
\hline
 & \multicolumn{2}{|c|}{TinyPerson V2} & \multicolumn{2}{|c}{TinyPerson} \\
\cline{2-5}
 & \#images & \#annotations & \#images & \#annotations \\
\hline\hline
train set  & 5711 & 262063 & 794  & 42197 \\
valid set  &  568 &  42399 & 816  & 30454 \\
test set   & 5753 & 315165 & -    & -\\
sum        & 12032& 619627 & 1610 & 72651 \\
\hline
\end{tabular}
}
\end{center}
\vspace{-4mm}
\caption{Statistic information of SeaPerson and TinyPerson.}
\label{tab: volumn of datasets}
\end{table}

\begin{table}
\begin{center}
\resizebox{0.48\textwidth}{!}{
\begin{tabular}{l|c|c|c}
\hline
dataset & absolute size & relative size & aspect ratio \\
\hline\hline
COCO       & 99.5 $\pm$ 107.5& 0.190 $\pm$ 0.203 & 1.213 $\pm$ 1.337\\
TinyPerson & 17.0 $\pm$ 16.9 & 0.011 $\pm$ 0.010 & 0.690 $\pm$ 0.422\\
TinyPerson V2  & 22.6 $\pm$ 10.8 & 0.016 $\pm$ 0.007 & 0.723 $\pm$ 0.424\\
\hline
\end{tabular}
}
\end{center}
\vspace{-4mm}
\caption{The mean and standard deviation of absolute size, relative size
and aspect ratio of objects in different datasets. Size is defined as the square root of the product of width and height and aspect ratio is the value of width divided by height. Absolute size is the size of object and relative size is the value of the object's size divided by the image's size. These settings are followed as~\cite{Yu2020ScaleMF}.}
\label{tab: size of datasets}
\end{table}

\section{Details of Semantic Variance}

The $Var(x')$ and $Var(y')$ in Eq.~\ref{Eq:a semantic variance} are calculate as:
\begin{equation}\small
\begin{aligned}
& Mean(x') = \frac{1}{M} \sum\limits_{1\leq j\leq M}{x'_j}; \\
& Mean(y') = \frac{1}{M} \sum\limits_{1\leq j\leq M}{y'_j}; \\
& Var(x') = \frac{1}{M}(x'_j - Mean(x'))^2;\\
& Var(y') = \frac{1}{M}(y'_j - Mean(y'))^2; 
\label{Eq: semantic variance detail}
\end{aligned}
\end{equation}

\noindent where $(x'_j, y'_j)$ is relative position of $j$-th object, $M$ is the number of objects in dataset.
For the objects whose annotated points or refined points are out of the bounding box, they often are regarded as wild points during the learning procedure. The wild points account for a small proportion and will not be learned by the network. However, their $RSV$ will be very large since the $RSV$ is relative to the height and width. Therefore, to better reflect the semantic variance of the points that really affect network learning, only the object whose annotated point or refined point is inside bounding box is used for calculating $RSV$.

\section{Visualization of CPR}

\begin{figure}[tb!]
  \centering
\includegraphics[width=8cm, height=5cm]{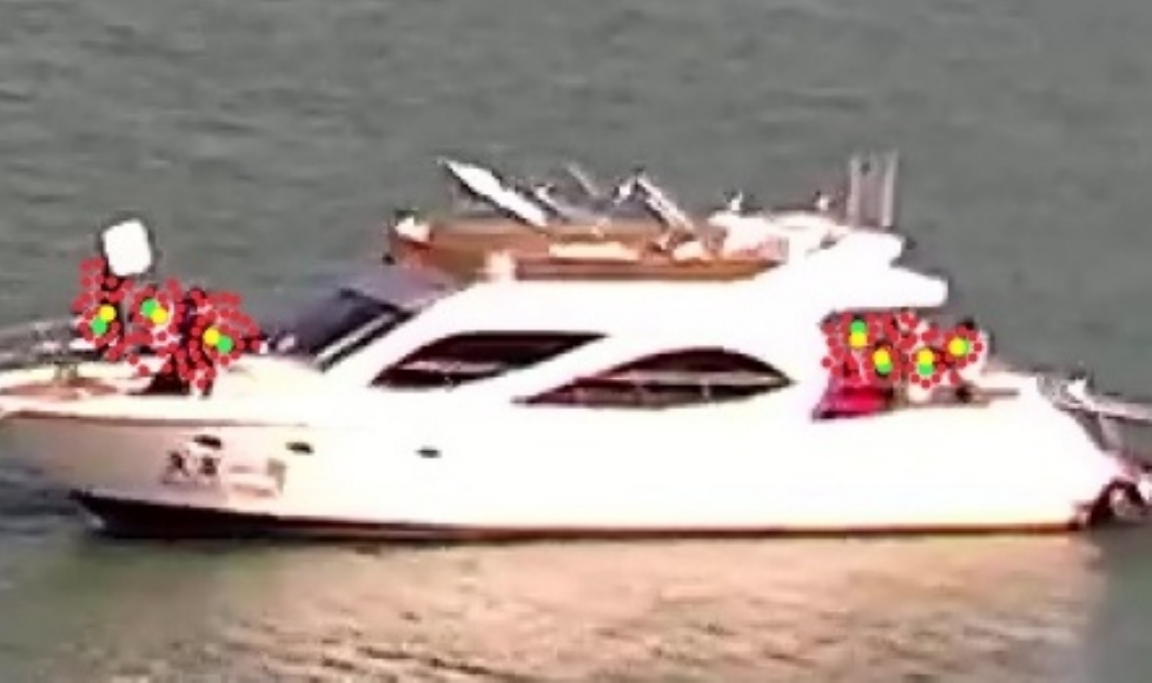}
\includegraphics[width=8cm, height=5cm]{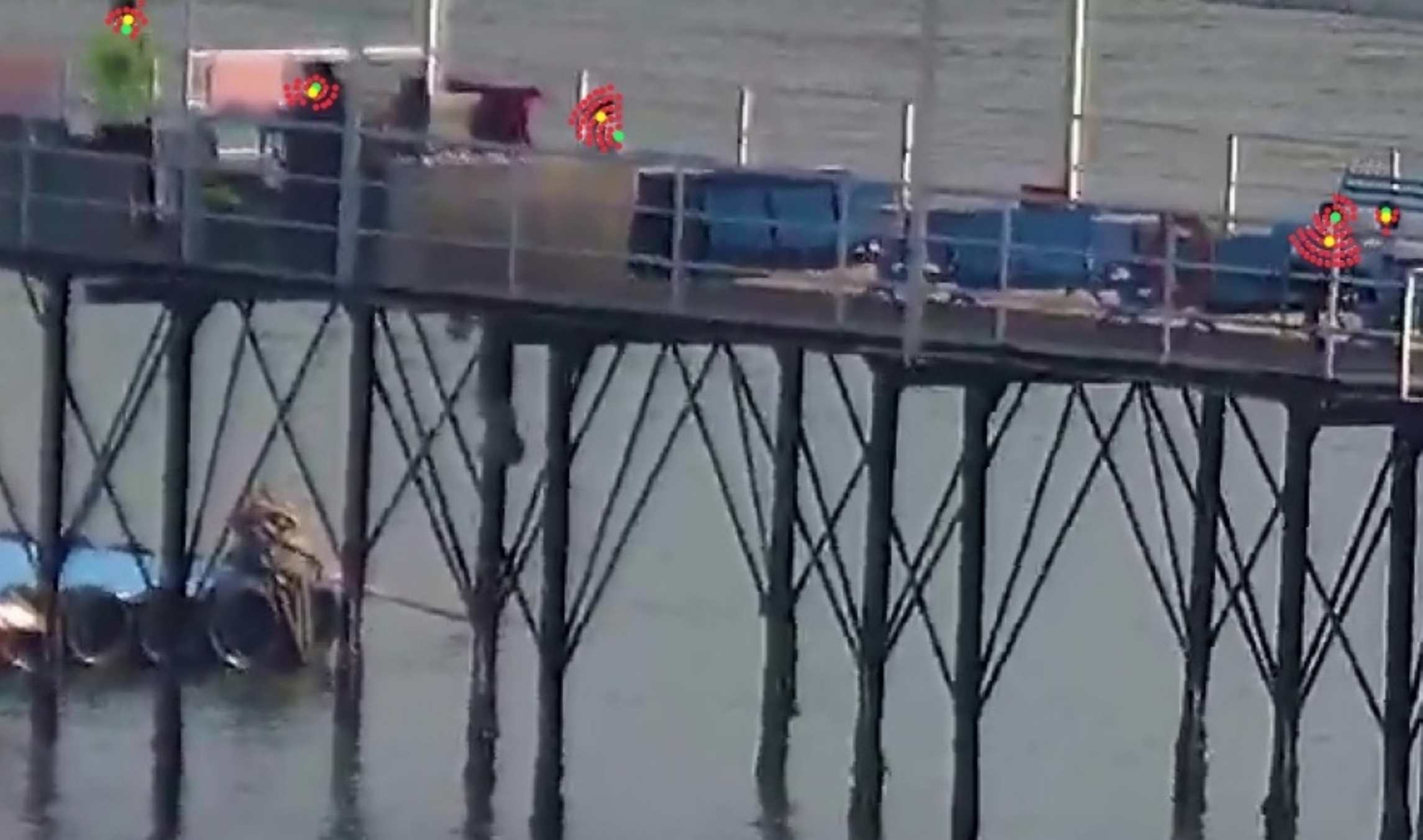}
\includegraphics[width=8cm, height=5cm]{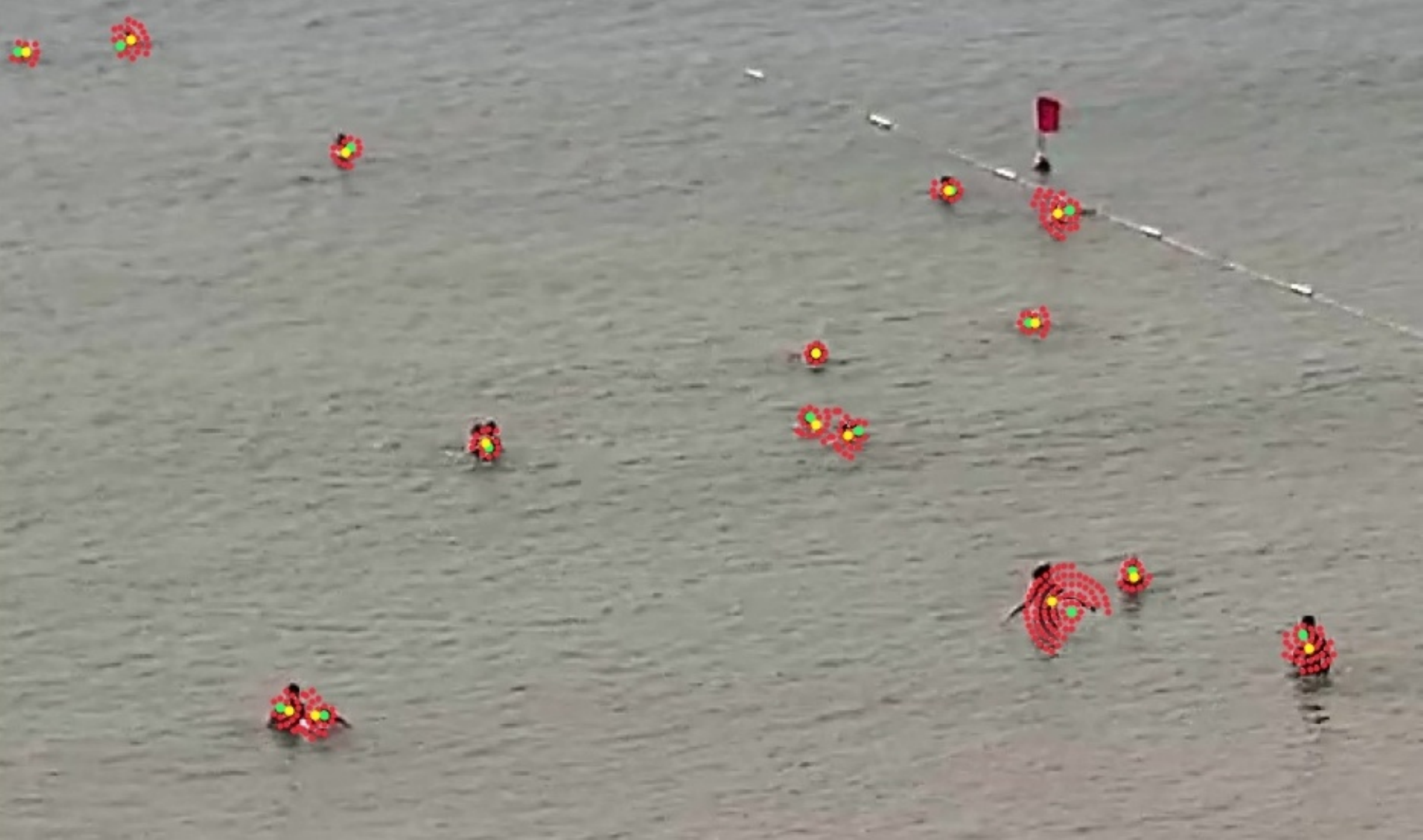}
\includegraphics[width=8cm, height=5cm]{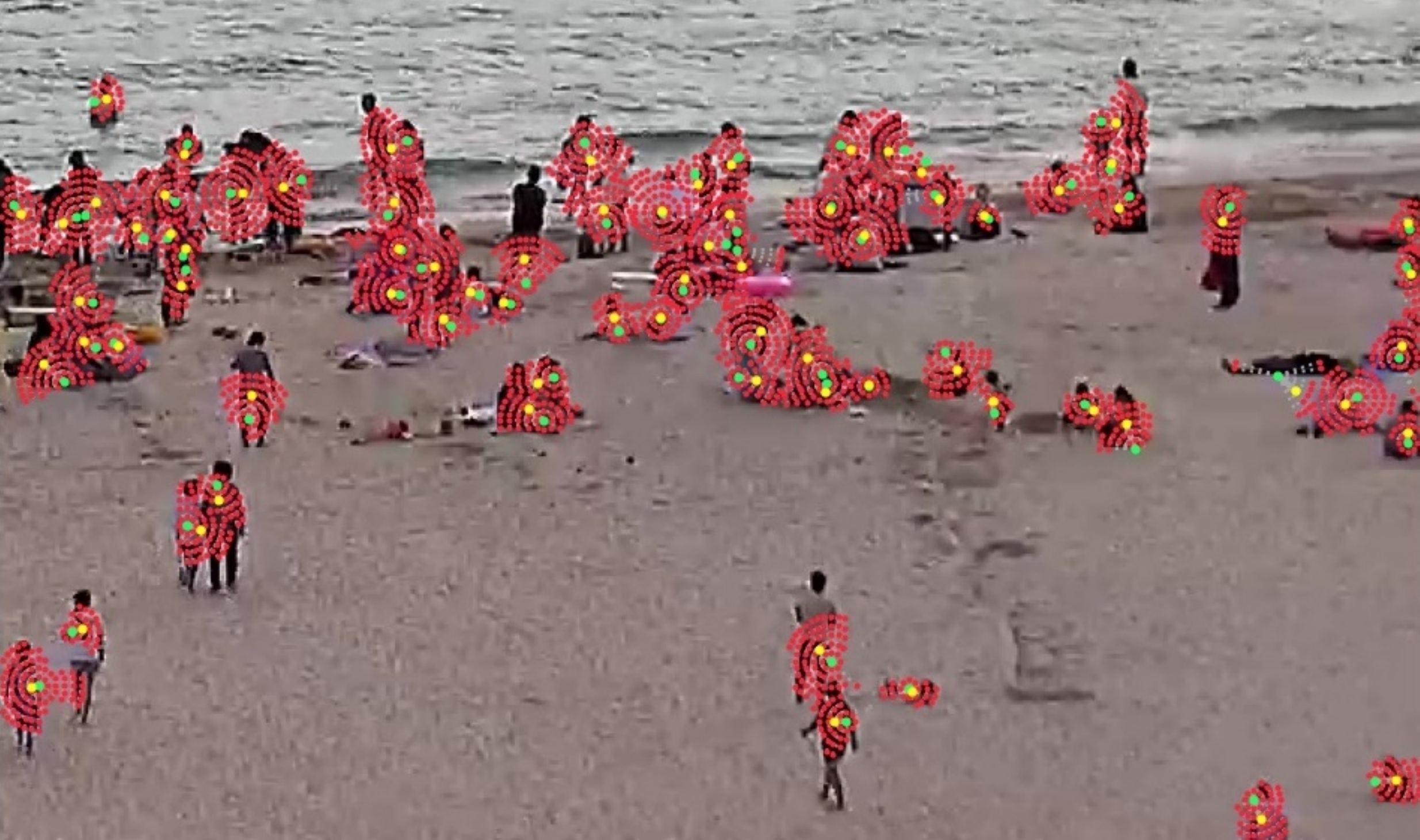}
\setlength{\belowcaptionskip}{-0.5cm}
  \caption{Visualization of CPR on SeaPerson. The images are cut from original images for better visualization. 
  Semantic points (red) around the annotated point (green) are weighted averaged to obtain the semantic center (yellow) as final refined point (see Sec.~\ref{sec: Refinement}).}
\label{fig:visulaization2 of CPR - seaperson}
\end{figure}

\begin{figure*}[tb!]
  \centering
\includegraphics[width=8.5cm, height=5cm]{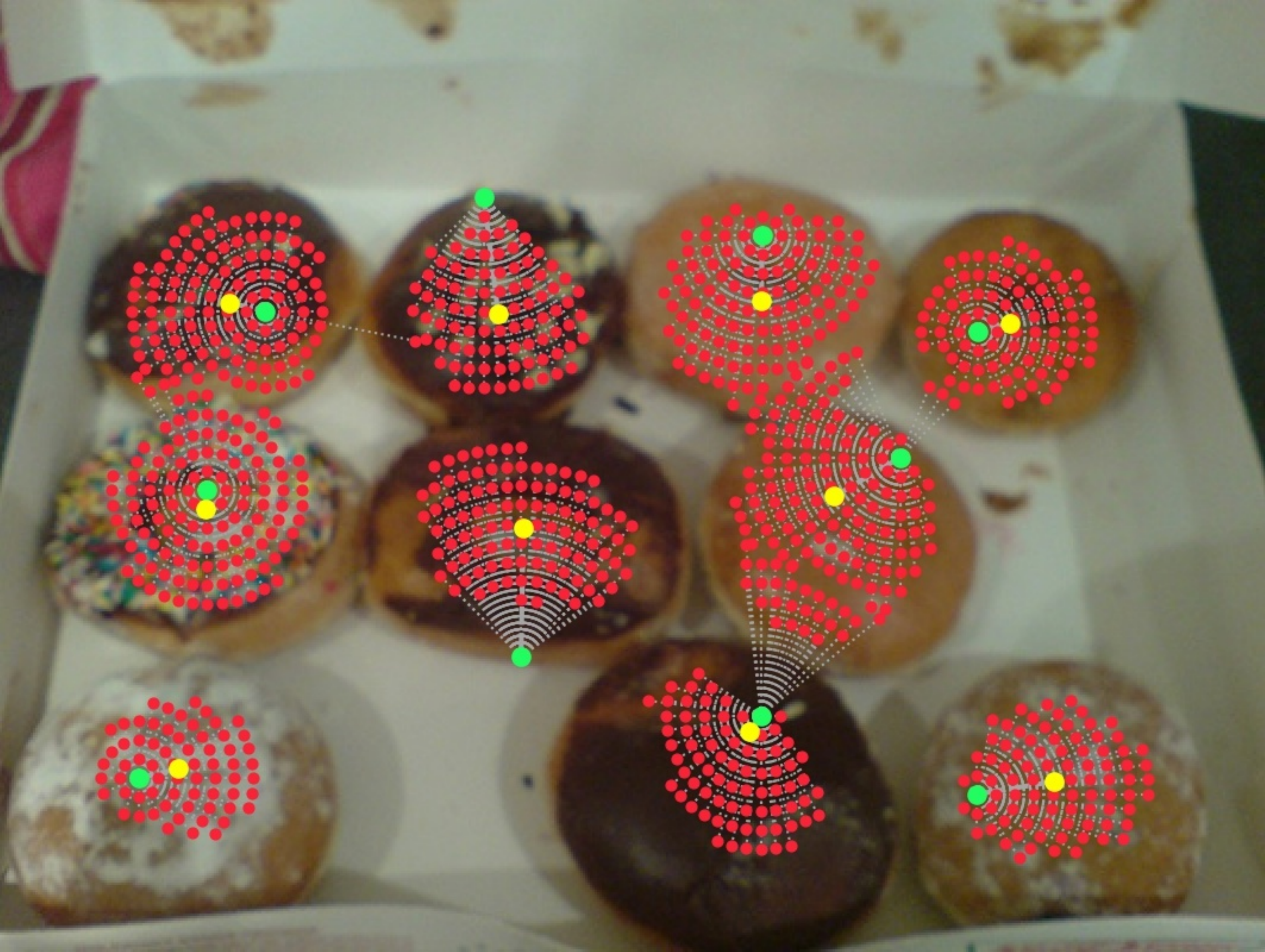}
\includegraphics[width=8.5cm, height=5cm]{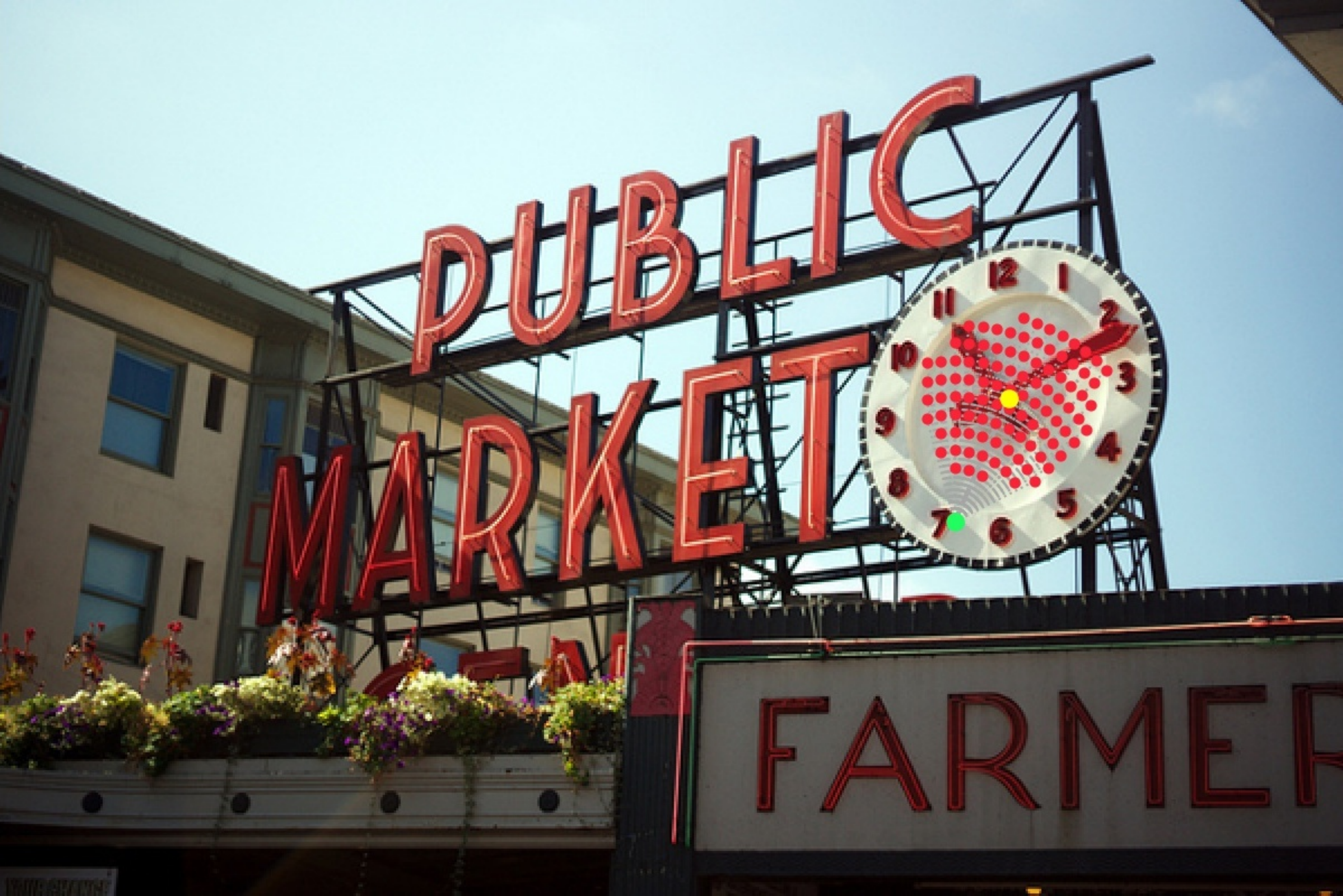}
\includegraphics[width=8.5cm, height=5cm]{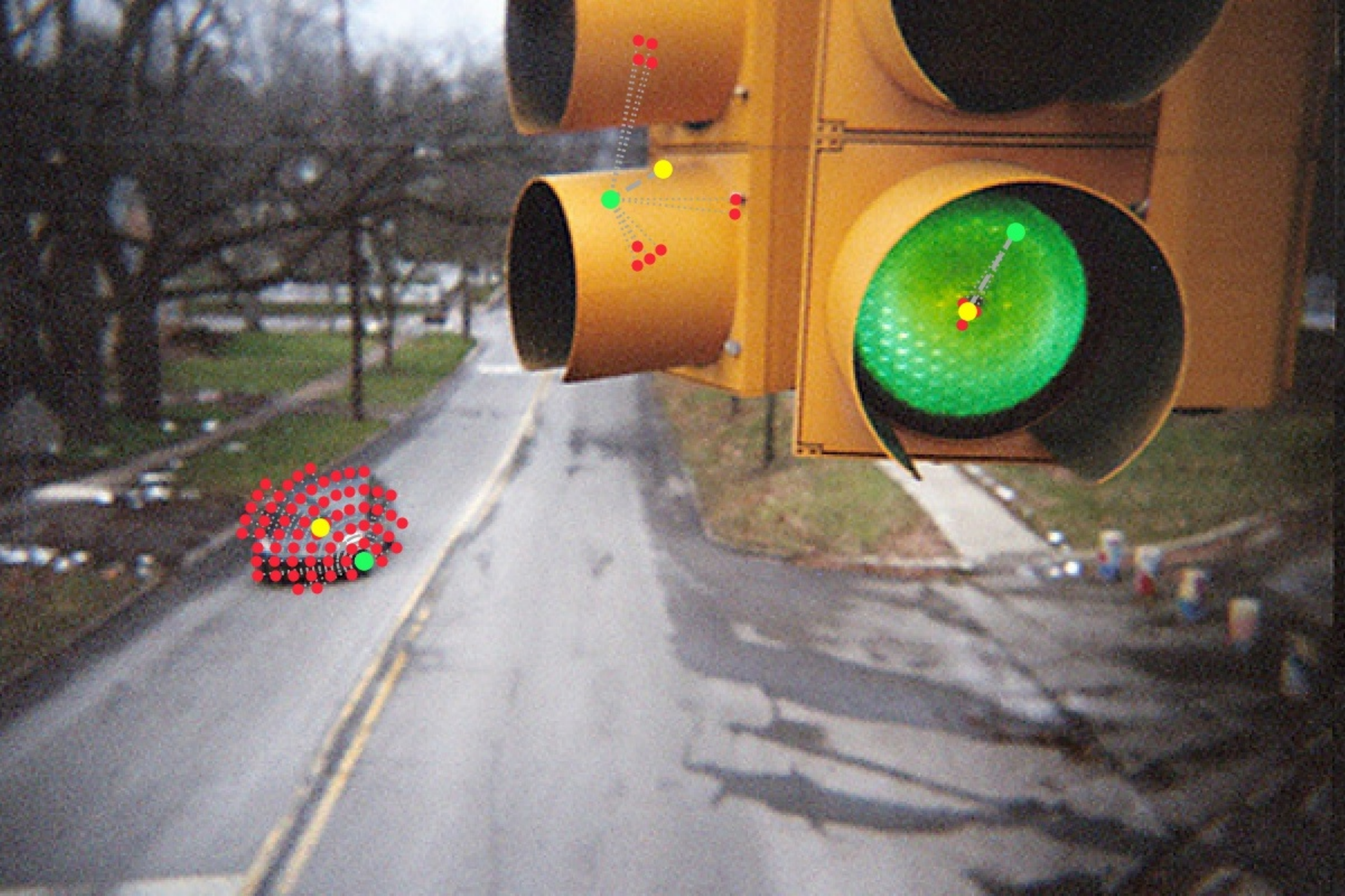}
\includegraphics[width=8.5cm, height=5cm]{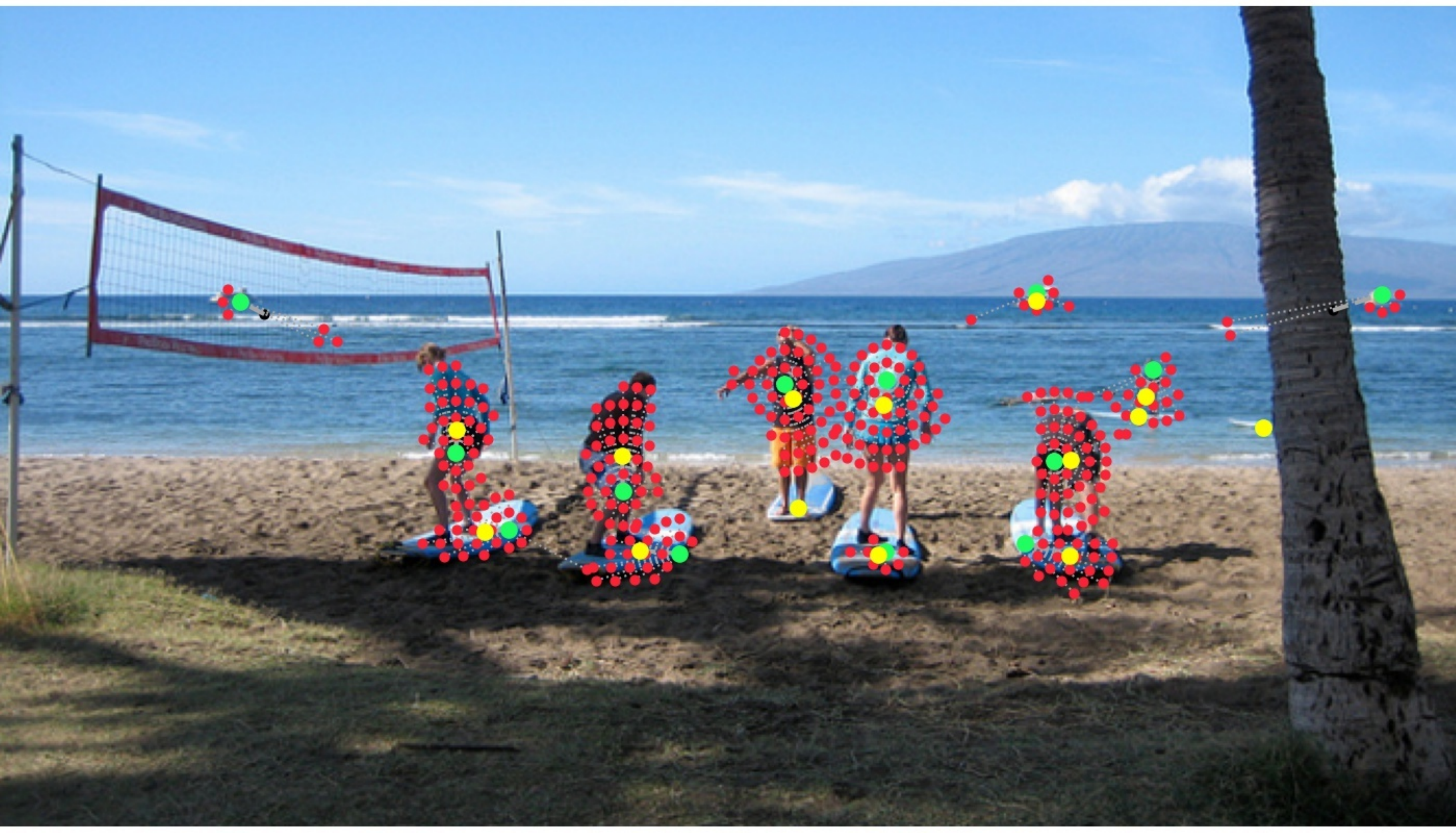}
\includegraphics[width=8.5cm, height=5cm]{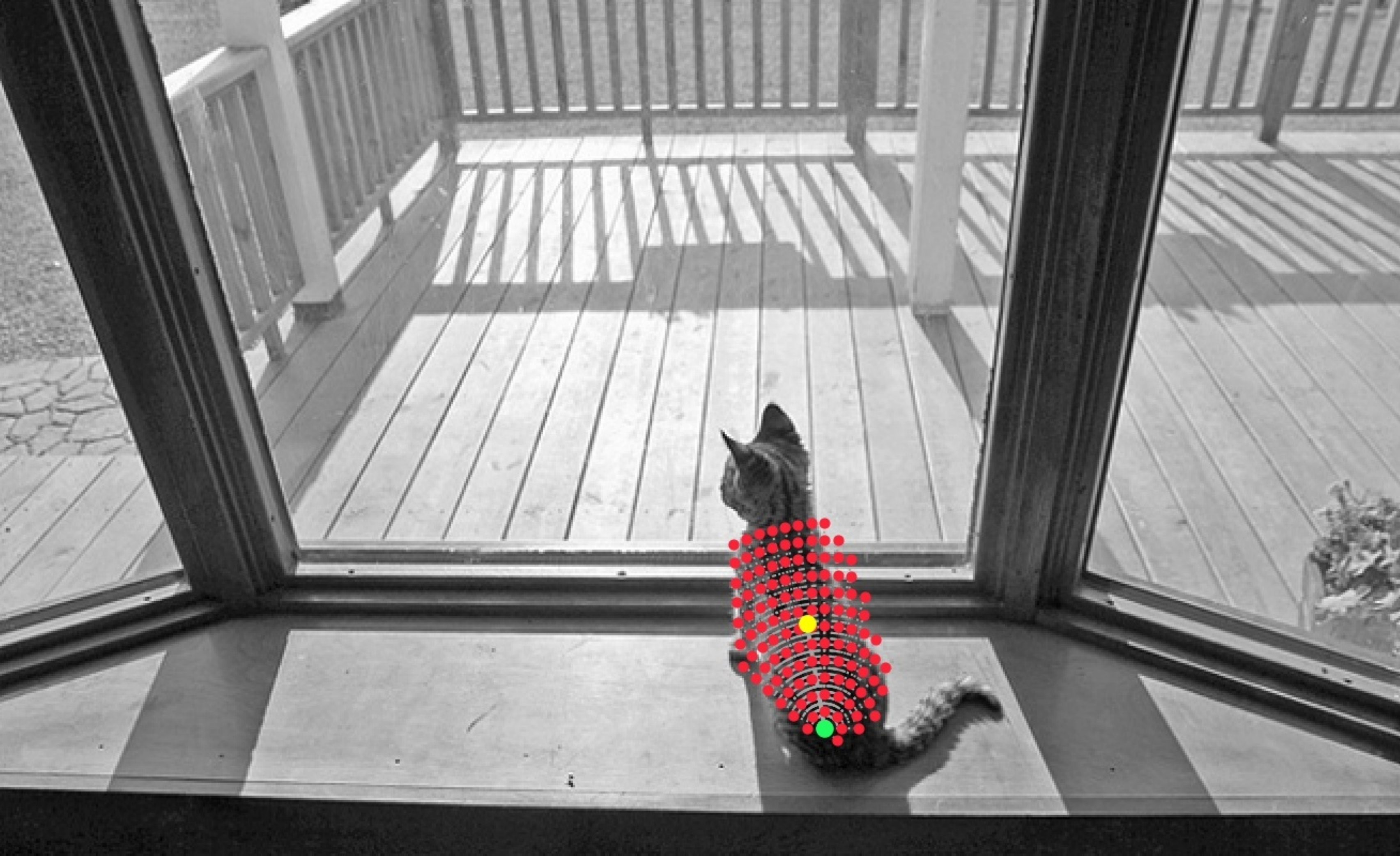}
\includegraphics[width=8.5cm, height=5cm]{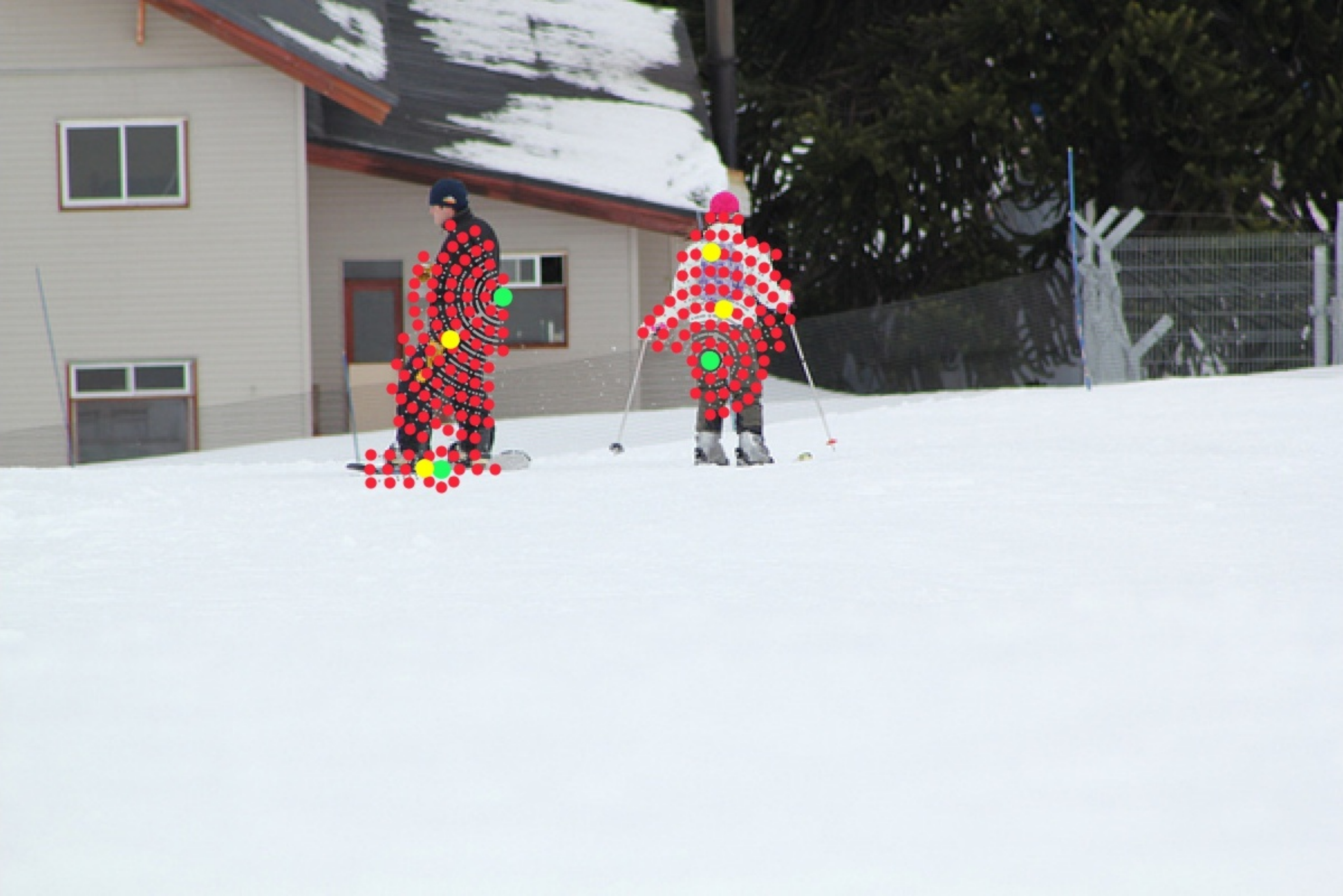}
\includegraphics[width=8.5cm, height=5cm]{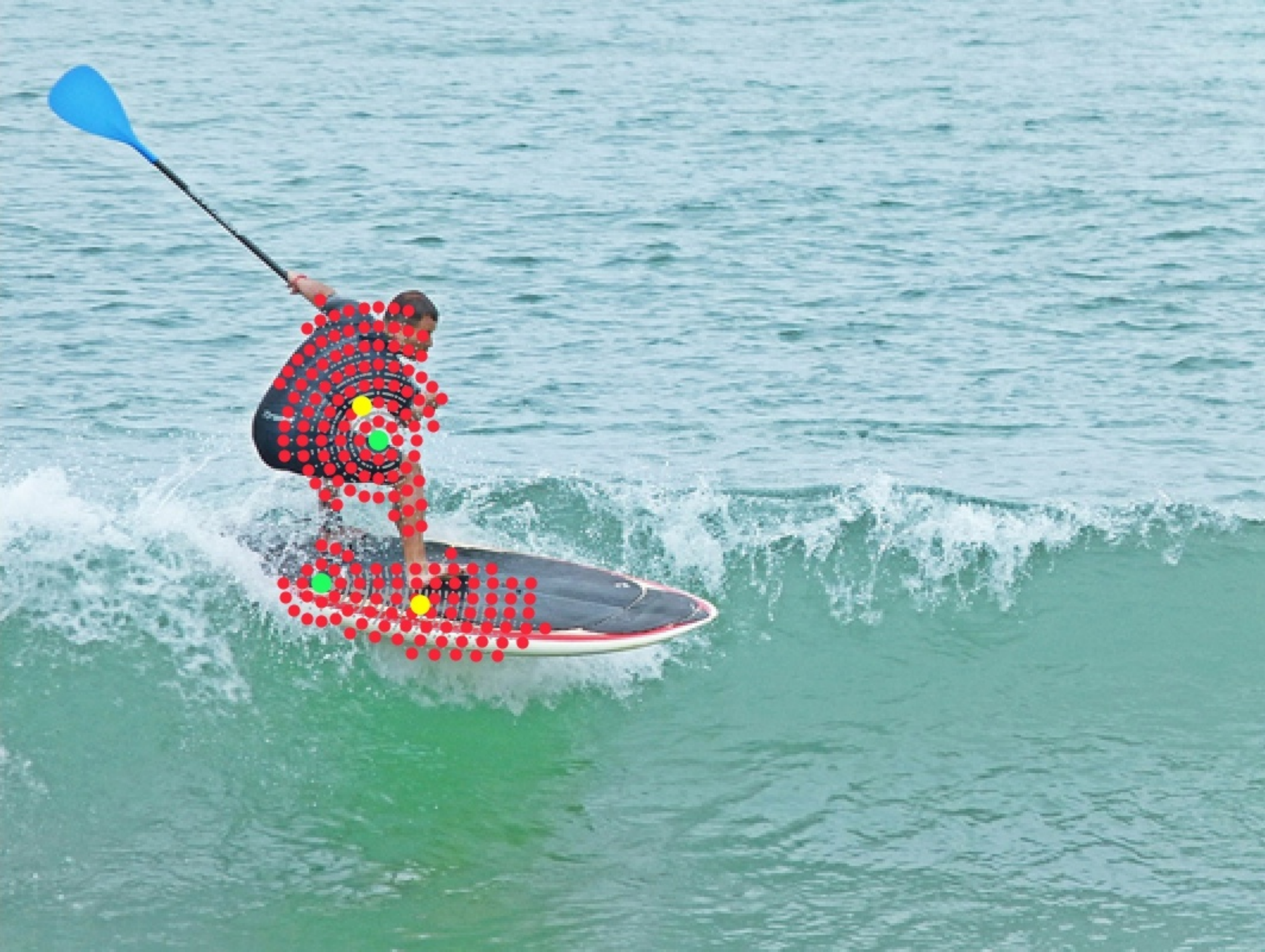}
\includegraphics[width=8.5cm, height=5cm]{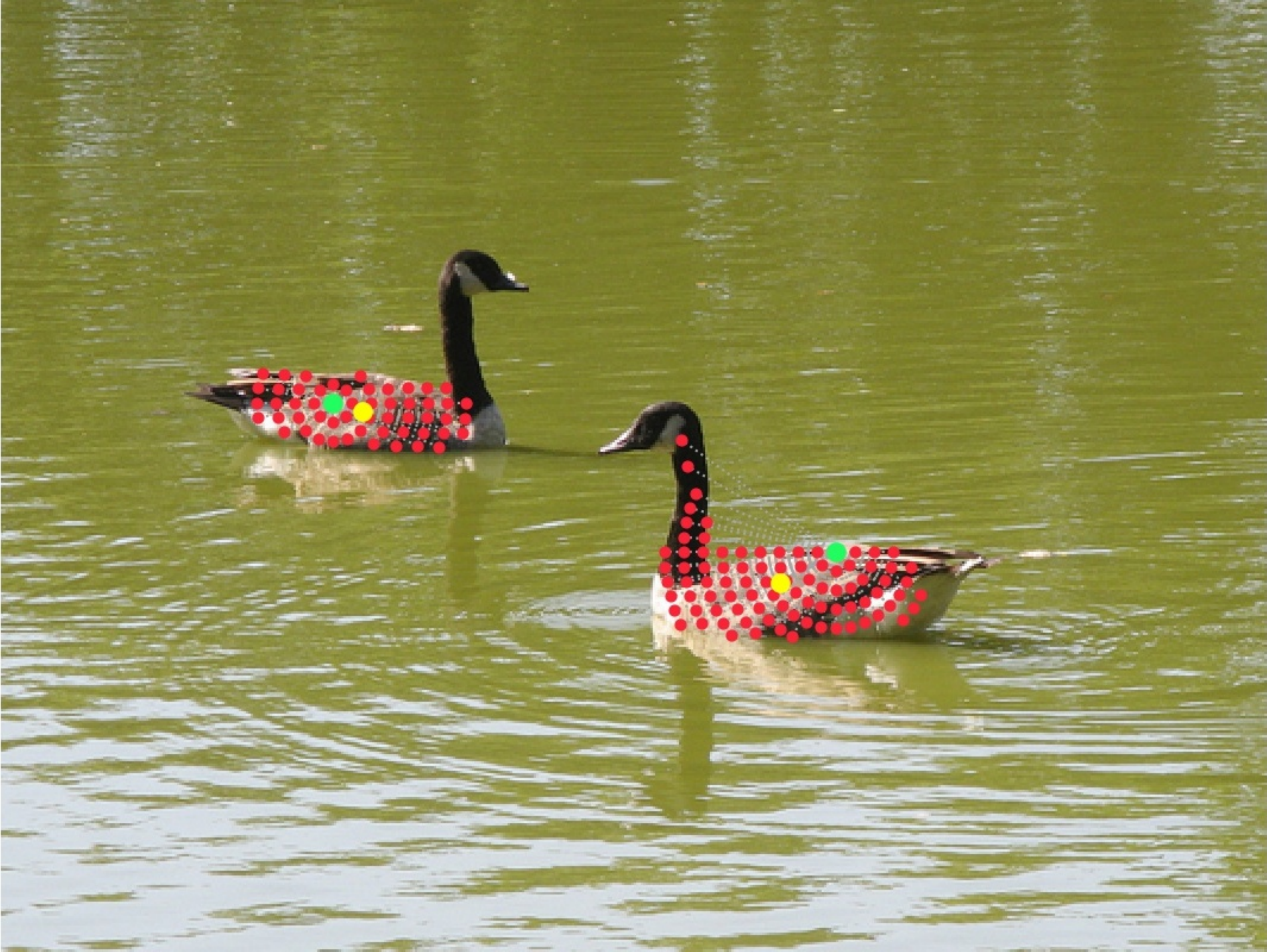}
\setlength{\belowcaptionskip}{-0.5cm}
  \caption{Visualization of CPR on COCO. The images are cut from original images for better visualization.
  Semantic points (red) around the annotated point (green) are weighted averaged to obtain the semantic center (yellow) as final refined point (see Sec.~\ref{sec: Refinement}).}
\label{fig:visulaization2 of CPR - coco}
\end{figure*}

\begin{figure*}[tb!]
  \centering
\includegraphics[width=8.5cm, height=5cm]{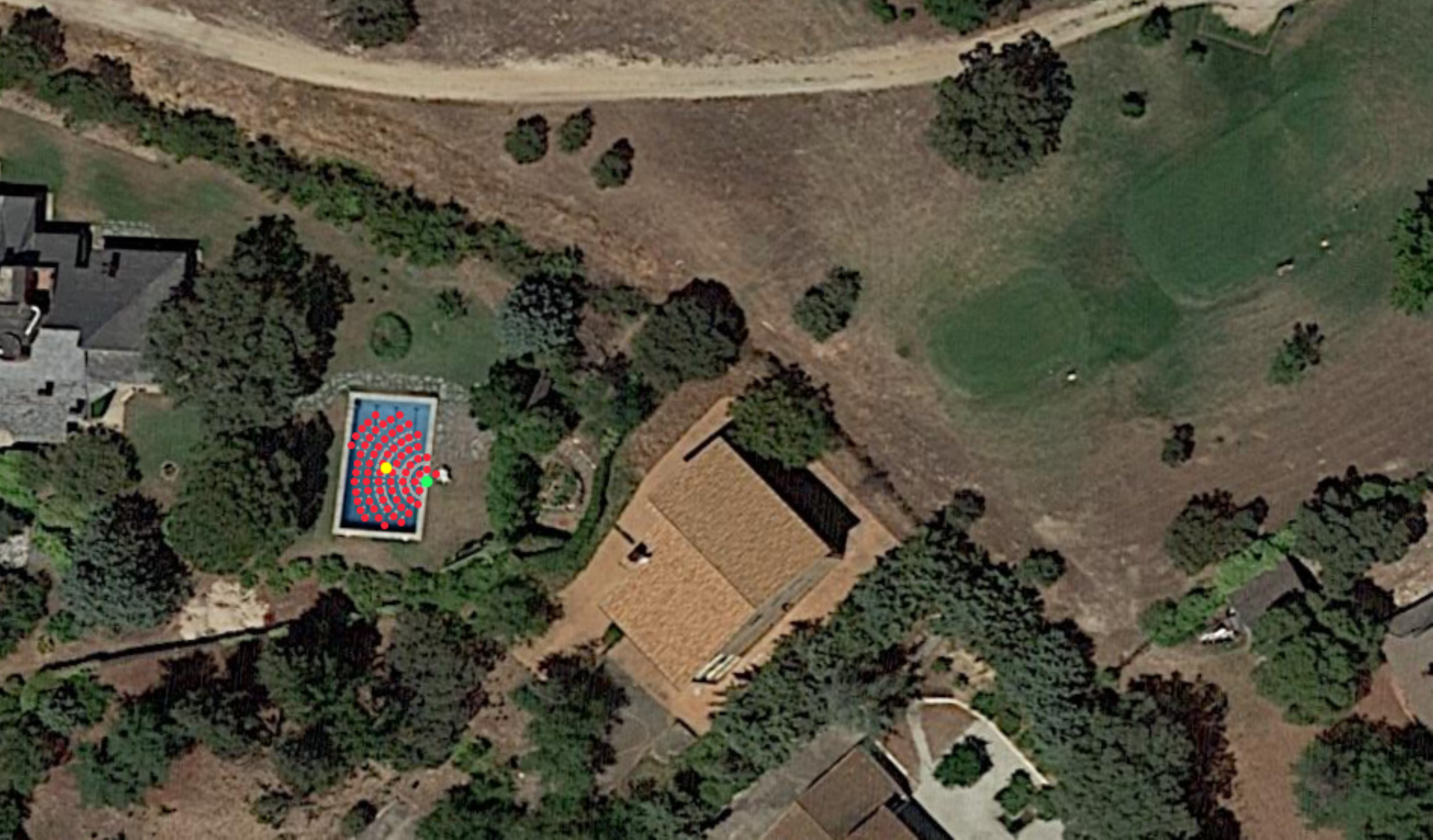}
\includegraphics[width=8.5cm, height=5cm]{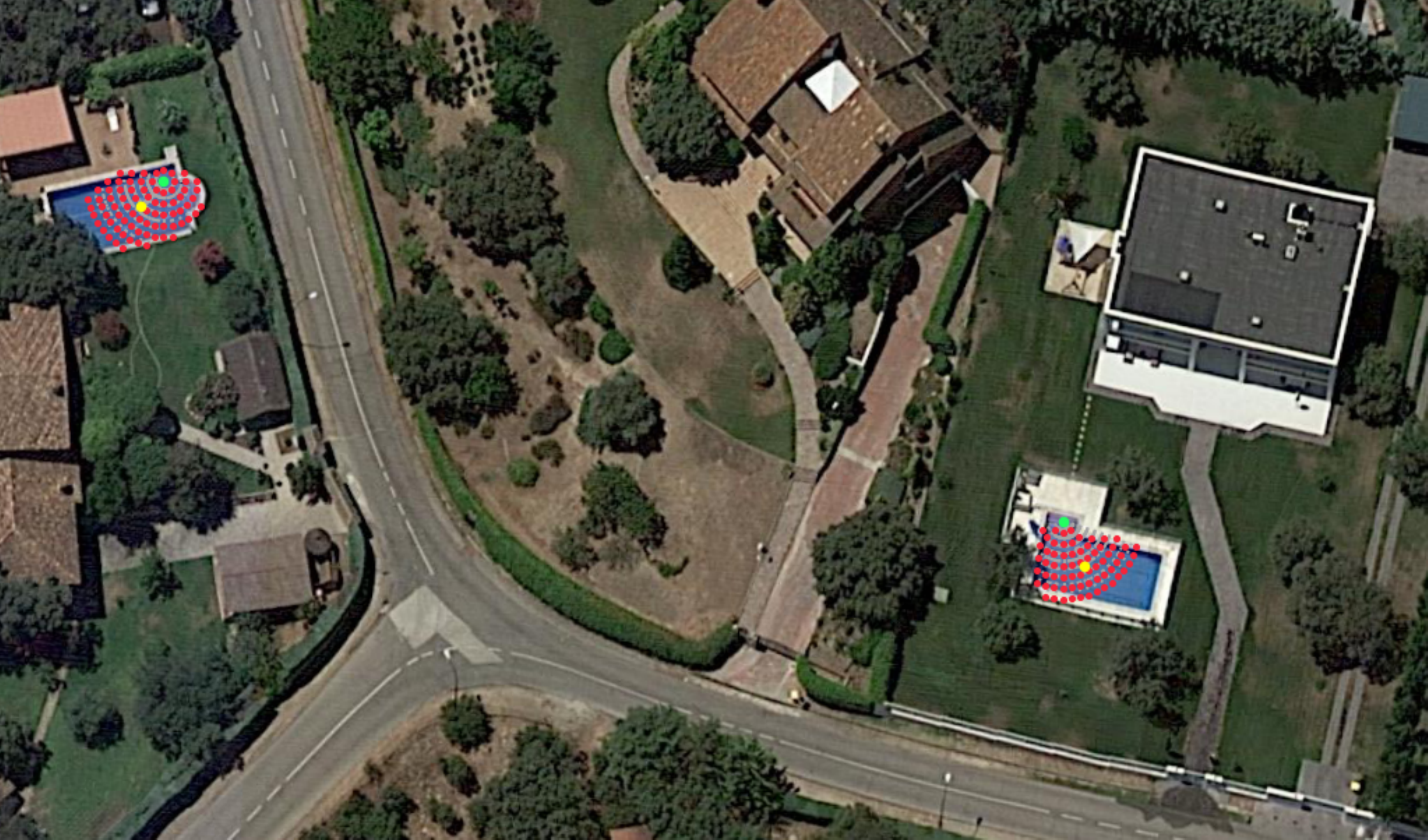}
\includegraphics[width=8.5cm, height=5cm]{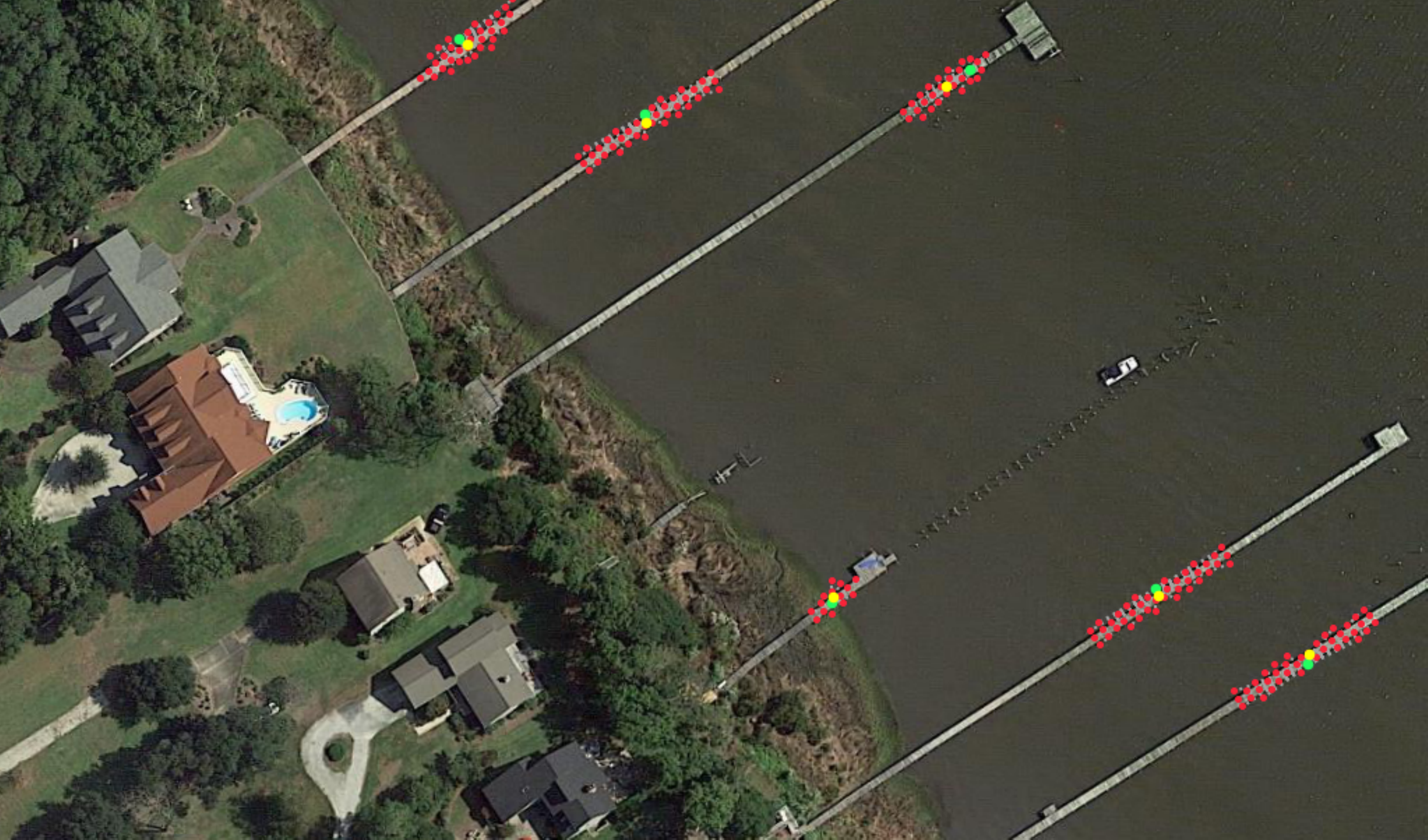}
\includegraphics[width=8.5cm, height=5cm]{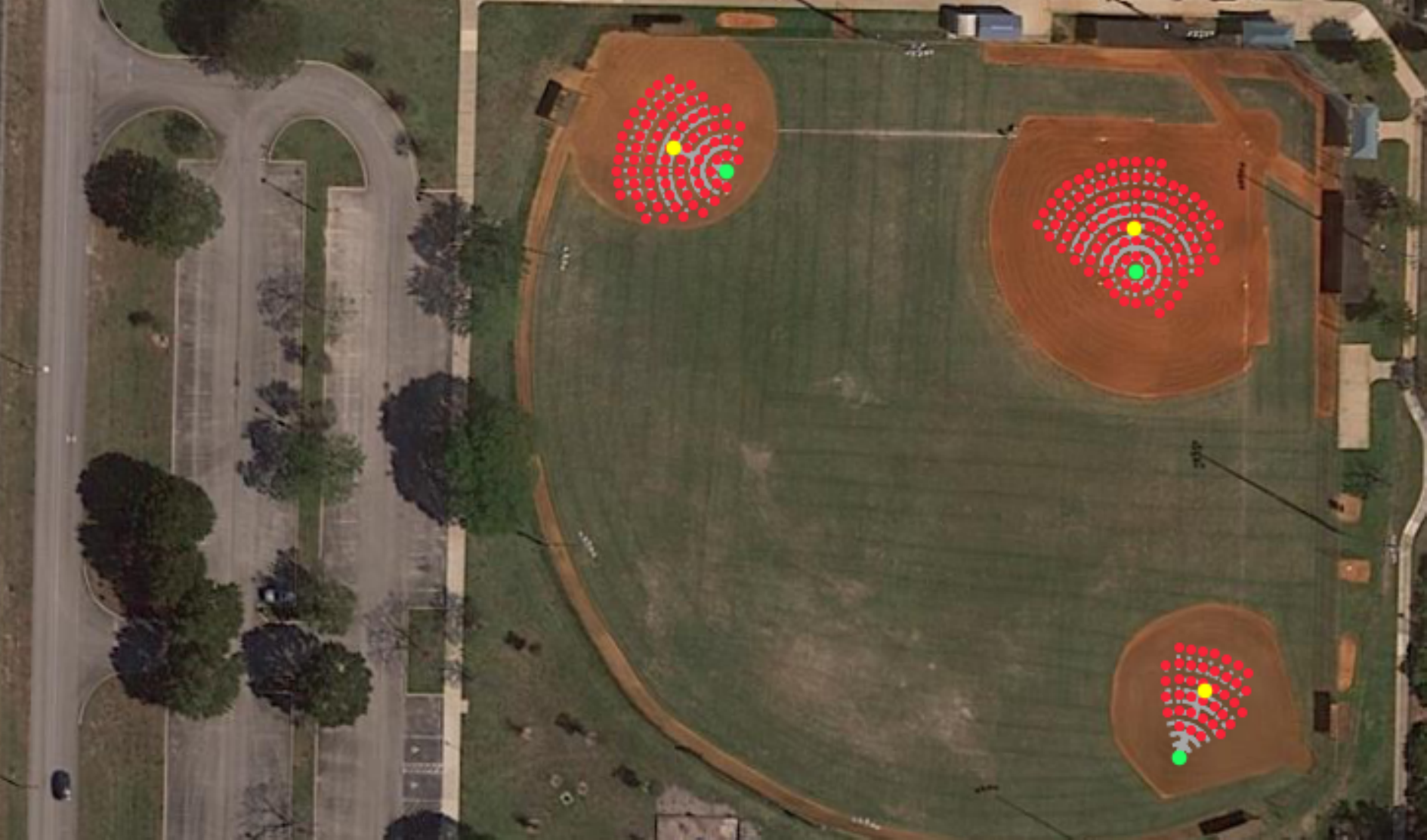}
\includegraphics[width=8.5cm, height=5cm]{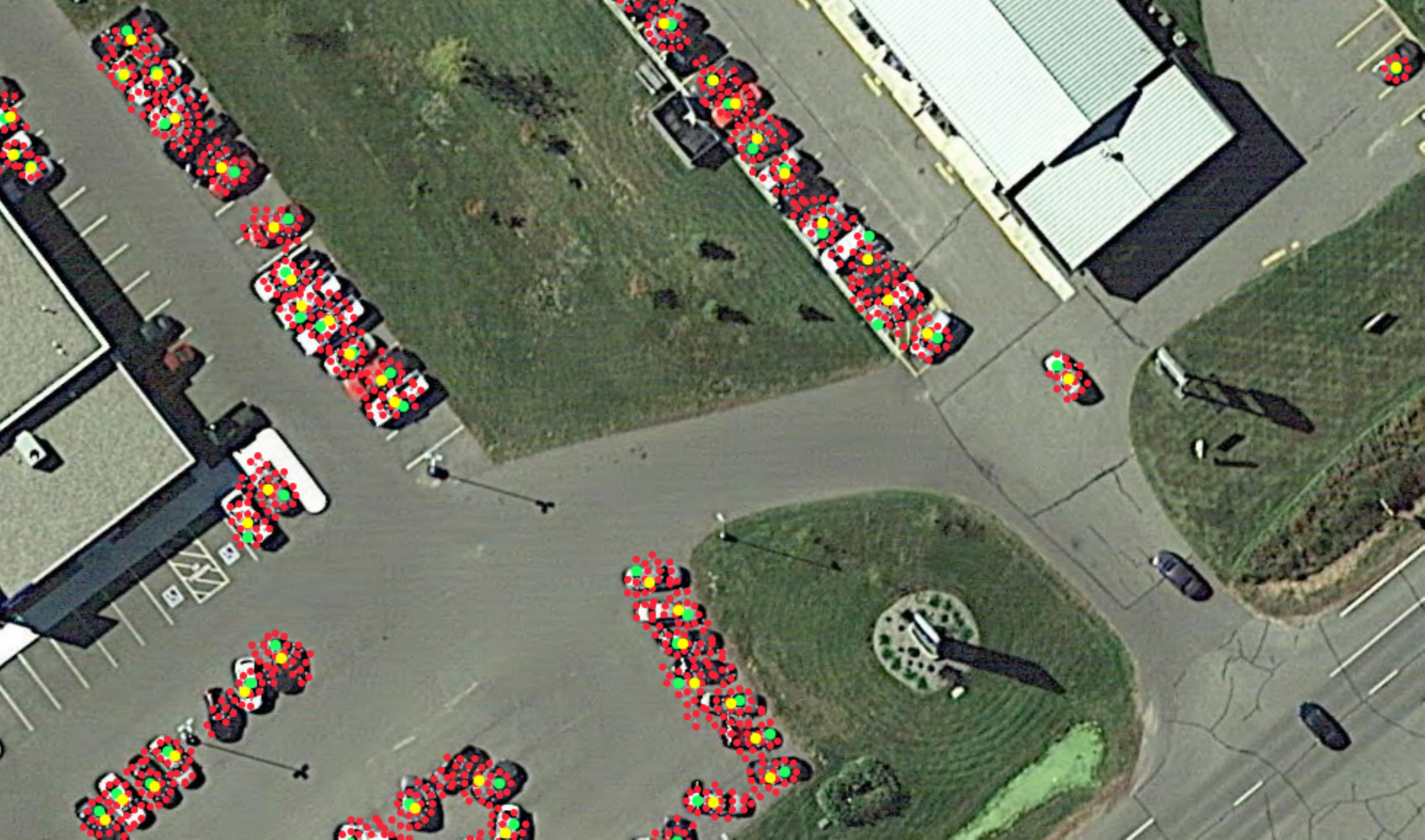}
\includegraphics[width=8.5cm, height=5cm]{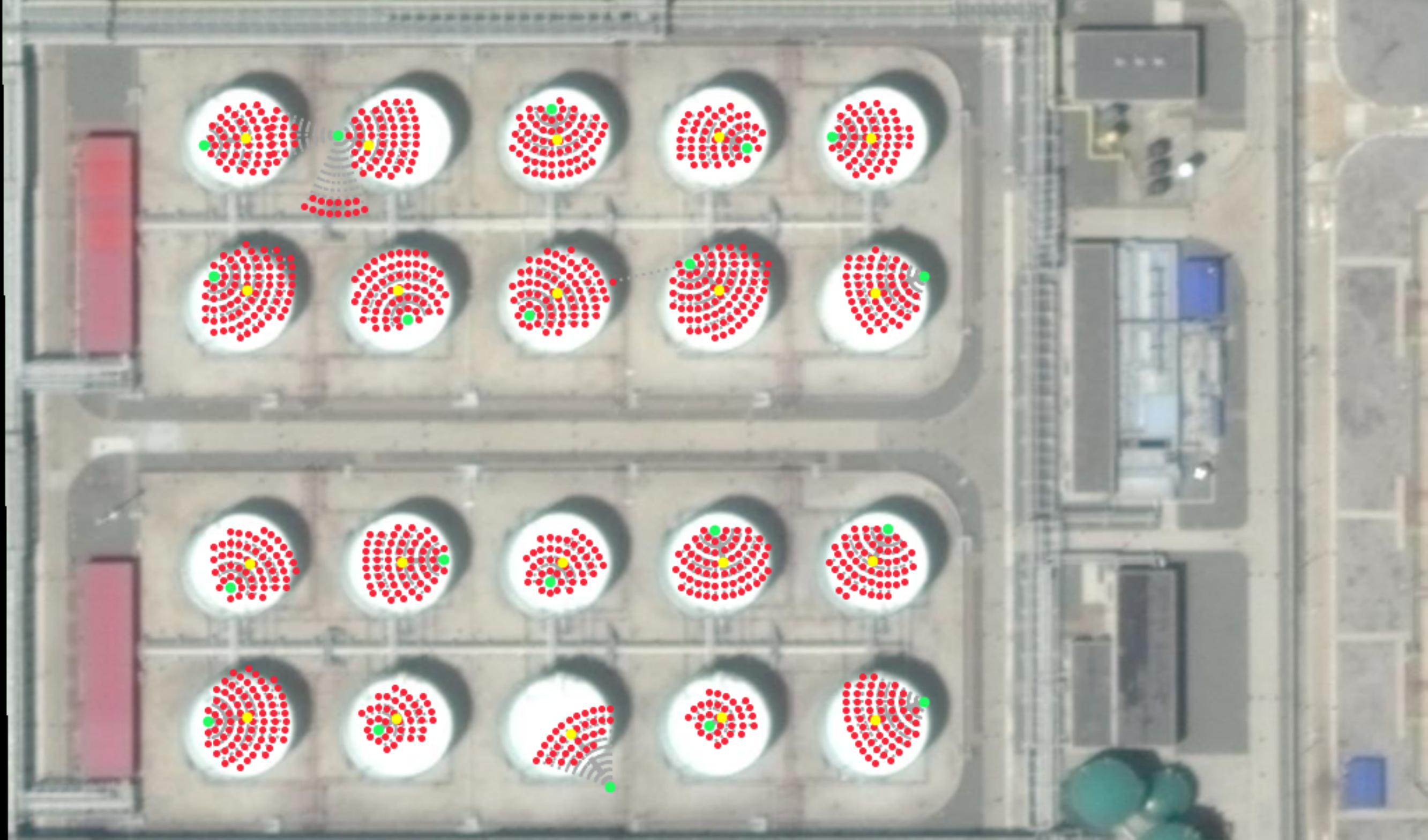}
\includegraphics[width=8.5cm, height=5cm]{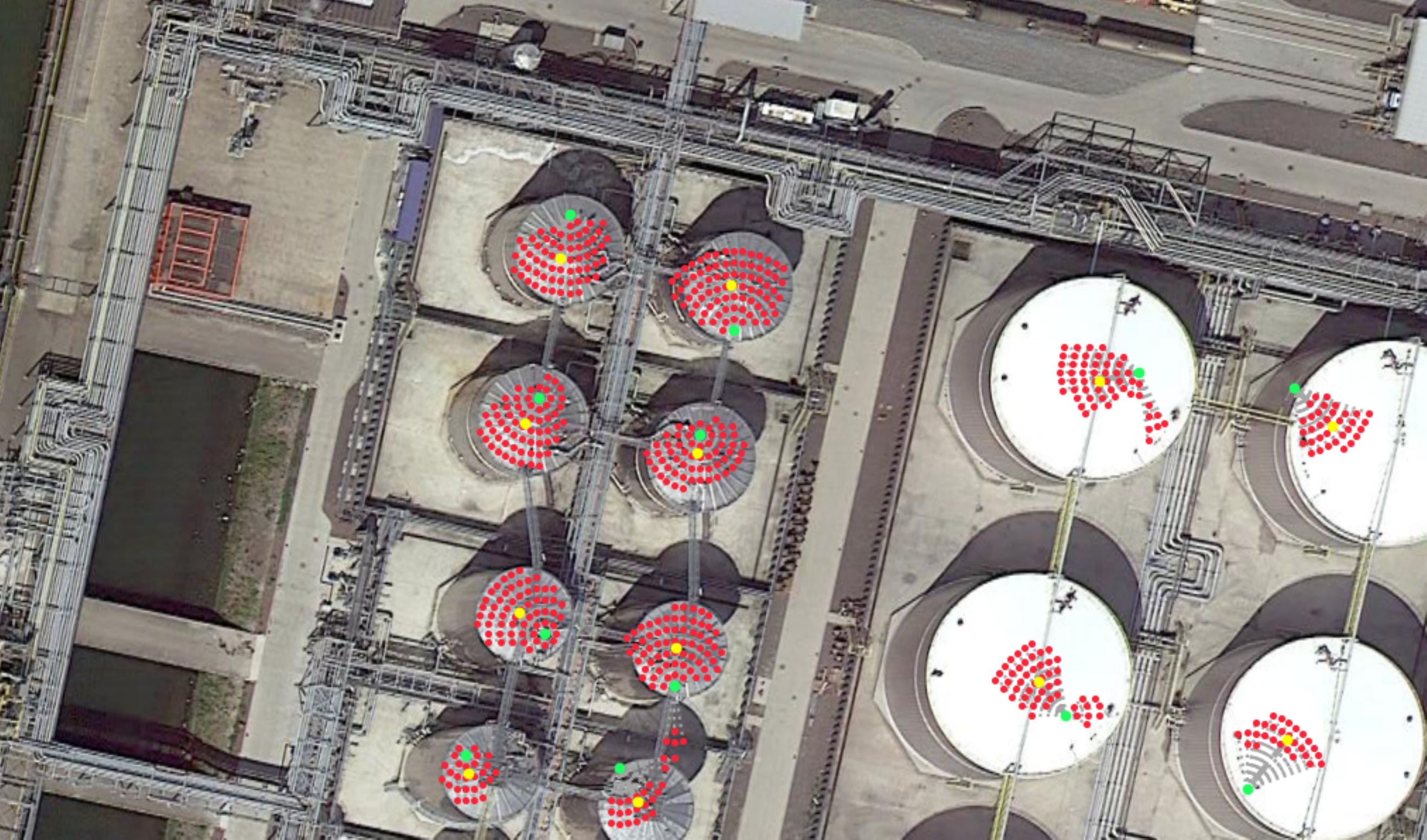}
\includegraphics[width=8.5cm, height=5cm]{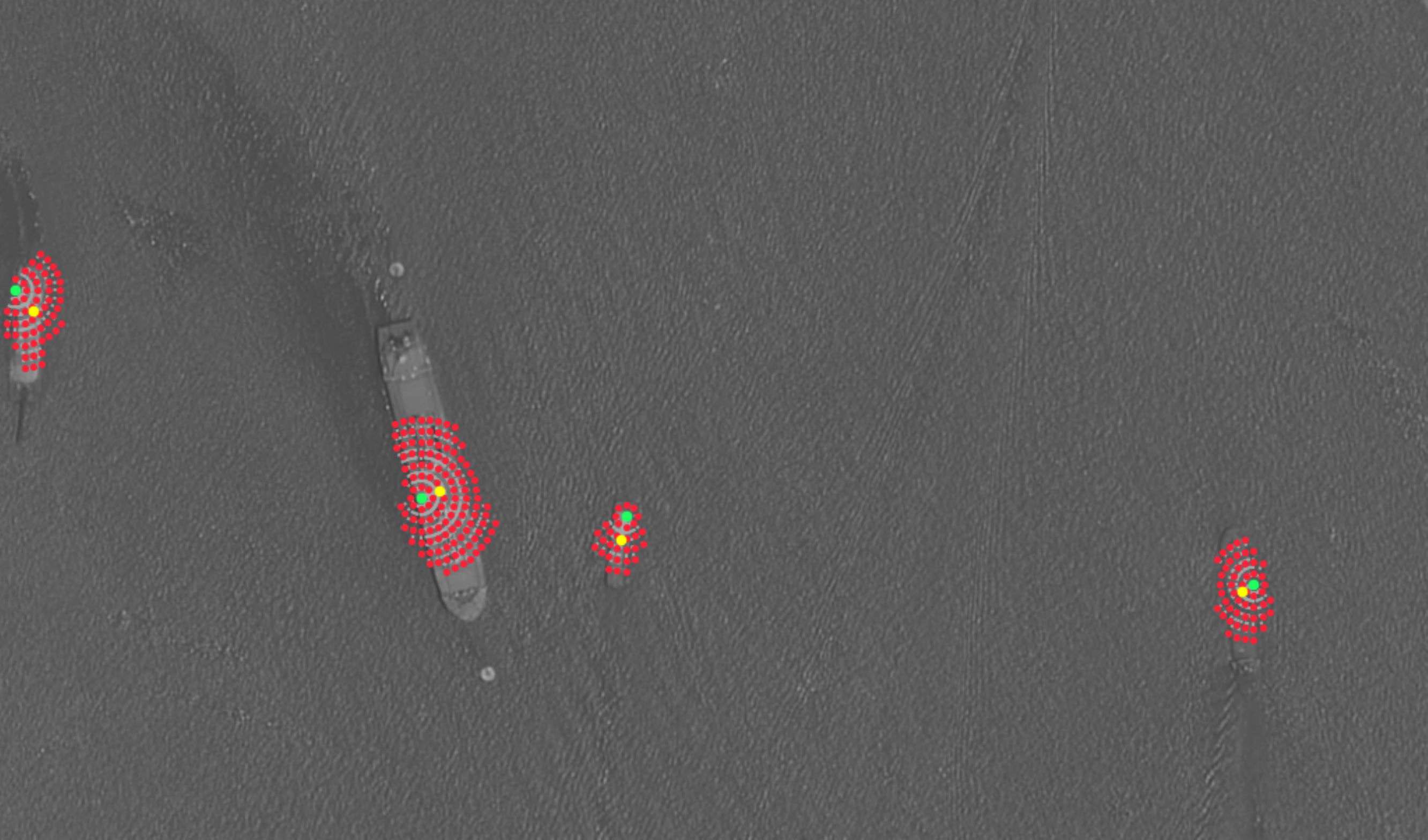}
\setlength{\belowcaptionskip}{-0.5cm}
  \caption{Visualization of CPR on DOTA. The images are cut from original images for better visualization.
  Semantic points (red) around the annotated point (green) are weighted averaged to obtain the semantic center (yellow) as final refined point (details in Sec.~\ref{sec: Refinement}).}
\label{fig:visulaization2 of CPR - dota}
\end{figure*}

The visulization of CPR on COCO, DOTA and SeaPerson are shown as Fig.~\ref{fig:visulaization2 of CPR - coco}, Fig.~\ref{fig:visulaization2 of CPR - dota} and Fig.~\ref{fig:visulaization2 of CPR - seaperson}, respectively.

\end{appendices}

\end{document}